\documentclass[conference]{IEEEtran}
\IEEEoverridecommandlockouts
\usepackage{cite}
\usepackage{amsmath,amssymb,amsfonts}
\usepackage{algorithmic}
\usepackage{graphicx}
\usepackage{textcomp}
\usepackage{cite}
\usepackage{xcolor}
\usepackage{hyperref}
\usepackage{subfigure}
\def\BibTeX{{\rm B\kern-.05em{\sc i\kern-.025em b}\kern-.08em
    T\kern-.1667em\lower.7ex\hbox{E}\kern-.125emX}}
\begin{document}

\title{Exploring the Connection Between Binary and Spiking Neural Networks\\
}

\author{\IEEEauthorblockN{Sen Lu, Abhronil Sengupta}
\IEEEauthorblockA{\textit{School of Electrical Engineering and Computer Science} \\
\textit{The Pennsylvania State University}\\
University Park, PA 16802, USA \\
Email: sengupta@psu.edu}}

\maketitle

\begin{abstract}
On-chip edge intelligence has necessitated the exploration of algorithmic techniques to reduce the compute requirements of current machine learning frameworks. This work aims to bridge the recent algorithmic progress in training Binary Neural Networks and Spiking Neural Networks - both of which are driven by the same motivation and yet synergies between the two have not been fully explored. We show that training Spiking Neural Networks in the extreme quantization regime results in near full precision accuracies on large-scale datasets like CIFAR-$100$ and ImageNet. An important implication of this work is that Binary Spiking Neural Networks can be enabled by ``In-Memory" hardware accelerators catered for Binary Neural Networks without suffering any accuracy degradation due to binarization. We utilize standard training techniques for non-spiking networks to generate our spiking networks by conversion process and also perform an extensive empirical analysis and explore simple design-time and run-time optimization techniques for reducing inference latency of spiking networks (both for binary and full-precision models) by an order of magnitude over prior work. Our implementation source code and trained models are available at 
\href{https://github.com/NeuroCompLab-psu/SNN-Conversion}{(Link)}. \end{abstract}

\begin{IEEEkeywords}
Spiking Neural Networks, Binary Neural Networks, In-Memory Computing\end{IEEEkeywords}

\section{Introduction}
The explosive growth of edge devices such as mobile phones, wearables, smart sensors and robotic devices in the current Internet of Things (IoT) era has driven the research for the quest of machine learning platforms that are not only accurate but are also optimal from storage and compute requirements. On-device edge intelligence has become increasingly crucial with the advent of a plethora of applications that require real-time information processing with limited connectivity to cloud servers. Further, privacy concerns for data sharing with remote servers have also fueled the need for on-chip intelligence in resourced constrained, battery-life limited edge devices.

To address these challenges, a wide variety of works in the deep learning community have explored mechanisms for model compression like pruning \cite{alvarez2017compression,han2015learning}, efficient network architectures \cite{iandola2016squeezenet}, reduced precision/quantized networks \cite{hubara2017quantized}, among others. In this work, we primarily focus on ``Binary Neural Networks" (BNNs) - an extreme form of quantized networks where the neuron activations and synaptic weights are represented by binary values \cite{courbariaux2016binarized, rastegari2016xnor}. Recent experiments on large-scale datasets like ImageNet \cite{deng2009imagenet} have demonstrated acceptable accuracies of BNNs, thereby leading to their current popularity. For instance, Ref. \cite{rastegari2016xnor} has shown that $58\times$ reduction in computation time and $32\times$ reduction in model size can be achieved for a BNN over a corresponding full-precision model. The drastic reductions in computation time simply result from the fact that costly Multiply-Accumulate operations required in a standard deep network can be simplified to simple XNOR and Pop-Count Operations. While current commercial hardware \cite{cass2019taking} already supports fixed point precision (as low as 4 bits), algorithmic progress on BNNs have contributed to the recent wave of specialized ``In-Memory" BNN hardware accelerators using CMOS \cite{zhang2017memory,biswas2018conv} and post-CMOS technologies \cite{sun2018xnor} that are highly optimized for single-bit state representations.

As a completely parallel research thrust, neuromorphic computing researchers have long advocated for the exploration of ``brain-like" computational models that abstract neuron functionality as a binary output ``spike" train over time. The binary nature of neuron output can be exploited to design event-driven hardware that is able to demonstrate significantly low power consumption by exploiting event-driven computation and data communication \cite{deng2020rethinking}. IBM TrueNorth \cite{akopyan2015truenorth} and Intel Loihi \cite{davies2018loihi} are examples of recently developed neuromorphic chips. While the power advantages of neuromorphic computing have been apparent, it has been difficult to scale up the operation of such ``Spiking Neural Networks" (SNNs) to large-scale machine learning tasks. However, recent work has demonstrated competitive accuracies of SNNs in large-scale image recognition datasets like ImageNet by training a non-spiking deep network and subsequently converting it to a spiking version for event-driven inference \cite{sengupta2019going,rueckauer2017conversion}.

There has not been any exploration or empirical study at exploring whether SNNs can be trained with binary weights for large-scale machine learning tasks. Note that this is not a trivial task since training standard SNNs itself from non-spiking networks has been a challenge due to the several constraints imposed on the base non-spiking network \cite{sengupta2019going}. If we assume that, in principle, such a network can be trained then the underlying enabling hardware for both BNNs and SNNs become equivalent \footnote{``near-equivalent" since neuron states are discretized as $-1,+1$ in BNN while SNN neuron outputs are discretized as $0,1$} (due to the binary nature of neuron/synapse state representation) except for the fact that the SNN needs to be operated over a number of time-steps. This work is aimed at exploring this connection between BNN and SNN.

While a plethora of custom BNN hardware accelerators have been developed recently, it is well known that BNNs suffer from significant accuracy degradation in complex datasets in contrast to full-precision networks. Recent work has demonstrated that while weight binarization can be compensated by training the network with the weight discretization in-loop, neuron activation binarization is a serious concern \cite{hubara2017quantized}. Interestingly, it has been shown that although SNNs represent neuron outputs by binary values \cite{maass1997networks}, the information integration over time can be approximated as a Rectified Linear transfer function (which is the most popular neuron transfer function used currently in full-precision deep networks). Drawing inspiration from this fact, we explore whether SNNs can be trained with binary weights as a means to bridge the accuracy gap of BNNs. This opens up the possibility of using BNN hardware accelerators for resource constrained edge devices without compromising on the recognition accuracy. This work also serves as an important application domain for SNN neuromorphic algorithms that can be viewed as augmenting the computational power of current non-spiking binary deep networks.
\section{Related Work \& Main Contributions}
The obvious comparison point of this paper would be recent efforts at training quantized networks with bit-precision greater than single bit. There have been a multitude of approaches \cite{zhou2016dorefa,choi2018pact,zhang2018lq,zhou2017balanced,deng2018gxnor,li2016ternary} with recent efforts aimed at designing networks with hybrid precision where the bit-precision of each layer of the network can vary \cite{wu2018mixed,wang2019haq,chakraborty2019pca,prabhu2018hybrid}. However, in order to support variable bit-precision for each layer, the underlying hardware would need to be designed accordingly to handle mixed-precision (which usually is characterized by much higher area, latency and power consumption than BNN hardware accelerators. Further, peripheral circuit complexities like sense amplifier input offset, parasitics limit their scalability \cite{xue201924}). This work explores a complementary research domain where the core underlying hardware can be simply customized for a BNN. This enables us to leverage the recent hardware developments of ``In-Memory" BNN accelerators and provides motivation for the exploration of $time$ (SNN computing framework) rather than $space$ (Mixed Precision Neural Networks) as the information encoding medium to compensate for accuracy loss exhibited by BNNs. Distributing the computations over time also implies that the instantaneous power consumption of the network would be much lower than mixed-precision networks and approach that of a BNN in the worst-case (savings observed due to SNN event-driven behavior discussed in the next section) which is the critical parameter governing power-grid design and packaging cost for low-cost edge devices.

There has been also recent efforts by the neuromorphic hardware community at training SNNs for unsupervised learning with binary weights enabled by stochasticity of several emerging post-CMOS technologies \cite{suri2013bio,sengupta2018stochastic,srinivasan2019restocnet}. Earlier works on analog CMOS VLSI implementations of bistable synapses have been also explored \cite{indiveri2006vlsi}. However, such works have been typically limited to shallow networks for simple digit recognition frameworks and do not bear relevance to our current effort at training supervised deep BNNs/SNNs.

We utilize standard training techniques for non-spiking networks and utilize the trained models for conversion to a spiking network. We perform an extensive empirical analysis and substantiate several optimization techniques that can reduce the inference latency of spiking networks by an order of magnitude without compromising on the network accuracy. A key facet of our proposal is the run-time flexibility. Depending on the application level accuracy requirement, the network can be simply run for multiple time-steps while leveraging the core BNN-catered ``In-Memory" hardware accelerator.

\section{B-SNN Proposal}
We first review preliminaries of BNNs and SNNs from literature and subsequently describe our proposed B-SNN (SNN with binary weights).
\subsection{Binary Networks}
Our BNN implementation follows the XNOR-Net proposal in Ref. \cite{rastegari2016xnor}. While the feedforward dot-product is performed using binary values, BNNs maintain proxy full-precision weights for gradient calculation. To formalize, the dot-product computation between the full-precision weights and inputs is simplified in a BNN as follows:
\begin{equation}
    I*W\approx(\text{sign}(I)*\text{sign}(W)) \alpha
\end{equation}
where, $\alpha$ is a non-binary scaling factor determined by the L1-norm of the full-precision proxies \cite{rastegari2016xnor}. Straight-Through Estimator (STE) with gradient-clipping to ($-1,+1$) range is used during the training process \cite{rastegari2016xnor}. Note that the above formulation reduces both weights and neuron activations to $-1,+1$ values. Although a non-binary scaling factor is introduced per layer, yet the number of non-binary operations due to the scaling factor is significantly low.
\subsection{Spiking Networks} 
\begin{figure}
\centering
    \includegraphics[width=0.5\textwidth]{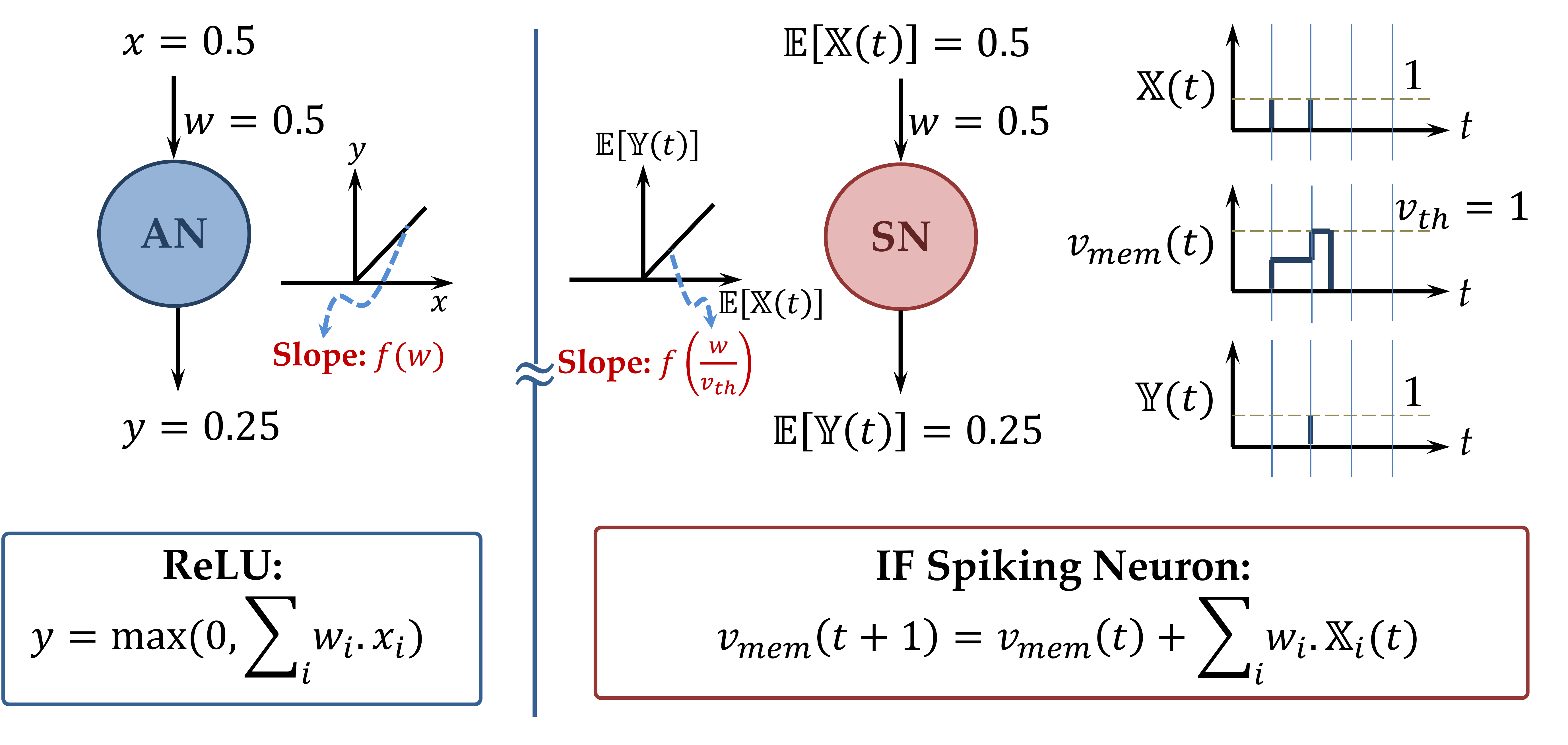}
\caption{An example to illustrate the mapping of ReLU to IF-Spiking Neuron.}
\label{fig:snn_example}
\end{figure}

SNN training can be mainly divided into three categories: ANN\footnote{ANN refers to standard non-spiking networks, Analog Neural Networks \cite{diehl2015fast}, where the neuron state representations are analog or full-precision in nature, instead of binary spikes.}-SNN conversion, backpropagation through time from scratch and unsupervised training through Spike-Timing Dependent Plasticity \cite{pfeiffer2018deep}. Since ANN-SNN conversion relies on standard backpropagation training techniques, it has been possible to scale SNN training using such conversion methods to large-scale problems \cite{sengupta2019going}. ANN-SNN conversion is driven by the observation that an Integrate-Fire spiking neuron is functionally equivalent to a Rectified Linear ANN neural transfer function. The functionality of an Integrate-Fire (IF) spiking neuron can be described by the temporal dynamics of a state variable, $v_{mem}$, that accumulates incoming spikes and fires an output spike whenever the membrane potential crosses a threshold, $v_{th}$.
\begin{equation}
v_{mem}(t+1) = v_{mem}(t) + \sum_{i} w_i.\mathbb{X}_i(t)
\label{lif}
\end{equation}
Considering $\mathbb{E}[\mathbb{X}(t)]$ to be the input firing rate (total spike count over a given number of time-steps), the output spiking rate of the neuron is given by $\mathbb{E}[\mathbb{Y}(t)] = \frac{w.\mathbb{E}[\mathbb{X}(t)]}{v_{th}}$ (considering the neuron being driven by a single input $\mathbb{X}(t)$ and a positive synaptic weight $w$). In case the synaptic weight is negative, the neuron firing rate would be zero since the neuron membrane potential would be unable to cross the threshold. This is in direct correspondence to the Rectified Linear functionality and is described by an example in Fig. \ref{fig:snn_example}. An ANN trained with ReLU neurons can therefore be transformed to an SNN with IF spiking neurons with minimal accuracy loss. The sparsity of binary neuron spiking behavior can be exploited for event-driven inference resulting in significant power savings \cite{sengupta2019going}.

\vspace{-6mm}
\subsection{Connecting Binary and Spiking Networks}
Our B-SNN is trained by using BNN training techniques described earlier. However, we utilize analog ReLU neurons instead of binary neurons. Conceptually, the network structure is analogous to Binary-Weight Networks (BWNs) introduced in Ref. \cite{rastegari2016xnor}. However, we also include additional constraints like bias-less neural units and no batch-normalization layers in the network structure \cite{sengupta2019going}. This is due to the fact that including bias and batch-normalization can potentially result in huge accuracy loss during the conversion process \cite{rueckauer2017conversion}. Much of the success of training BNNs can be attributed to Batch-Normalization. Hence, it is not trivial to train such highly-constrained ANNs with binary weights and without Batch-Normalization aiding the training process. Additional constraints like the choice of pooling mechanism, spiking neuron reset mechanism are discussed in details in the next section. This work is aimed at performing an extensive empirical analysis to substantiate the feasibility of achieving high-accuracy and low-latency B-SNNs. 

Note that the threshold of each network layer is an additional hyper-parameter introduced in the SNN model and serves as an important trade-off factor between SNN latency and accuracy. Due to the neuron reset mechanism, the SNN neurons are characterized by a discontinuity at the reset time-instants. If the threshold is too low, the membrane potential accumulations would be always higher than the threshold causing the neuron to continuously fire. On the other hand, too high thresholds result in increased latency for neurons to fire. In this work, we normalize the neuron thresholds to the maximum ANN activation (recorded by passing the training set once after the ANN has been trained) \cite{rueckauer2017conversion}. Other thresholding schemes can be also applied \cite{sengupta2019going} to minimize the conversion accuracy loss further.

\begin{figure}[htp]
  \centering
  \subfigure[BNN RRAM array]{\includegraphics[scale=0.55]{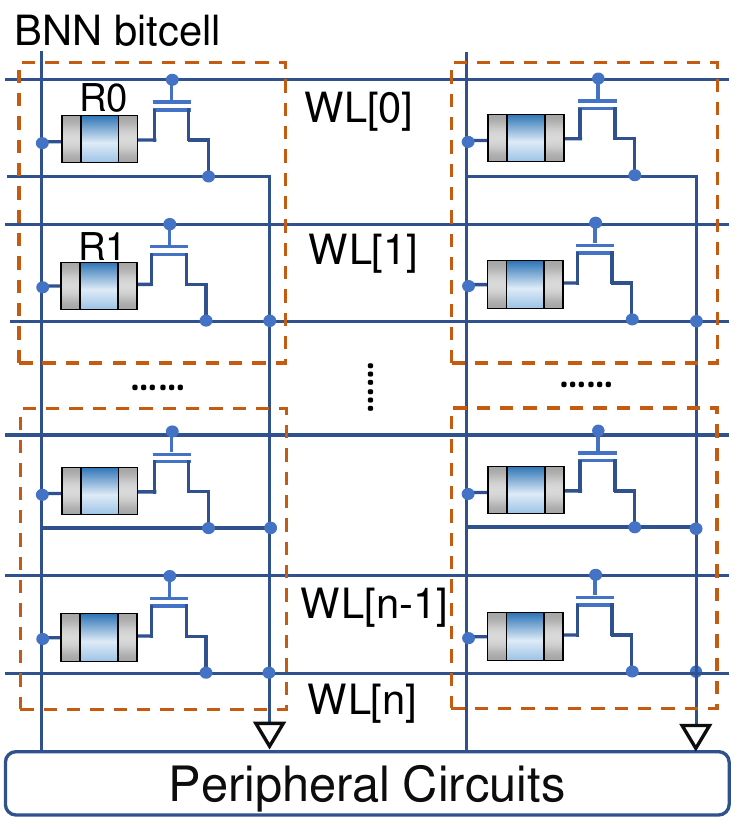}\label{fig:xnor_circuit}}\quad
  \subfigure[B-SNN RRAM array]{\includegraphics[scale=0.5]{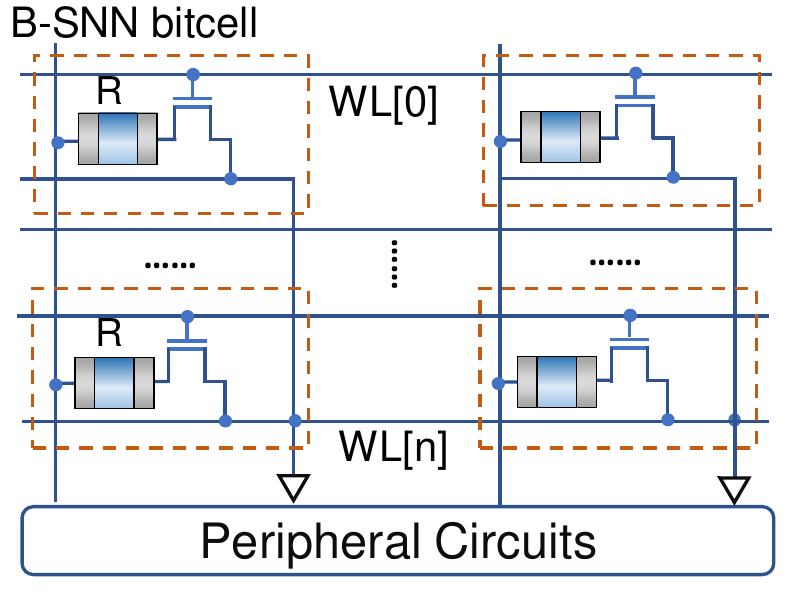}\label{fig:snn_circuit}}

\subfigure[BNN Input Encoding]{
    \begin{minipage}{1.5in}
    \centering
         \begin{tabular}{||c|c|c||} 
            \hline
            Input & WL[0] & WL[1] \\ 
            \hline\hline
            -1 & 1 & 0 \\ 
            \hline
            +1 & 0 & 1 \\
            \hline
         \end{tabular}  \label{tab:XNOR_input}
    \end{minipage}
}
\hfill 	
\subfigure[B-SNN Input Encoding]{
    \begin{minipage}{1.5in}
    \centering
        \begin{tabular}{||c|c||} 
            \hline
            Input & WL[0]  \\ 
            \hline\hline
            0 &  0 \\ 
            \hline
            1 & 1 \\
            \hline
        \end{tabular}  \label{tab:SNN_input}
    \end{minipage}
}

\subfigure[BNN Resistor State]{
    \begin{minipage}{1.5in}
    \centering
        \begin{tabular}{||c|c|c||} 
            \hline
            Weight & R0 & R1 \\ 
            \hline\hline
            -1 & LRS & HRS \\ 
            \hline
            +1 & HRS & LRS \\
            \hline
        \end{tabular}  \label{tab:XNOR_res}
    \end{minipage}
}
\hfill 	
\subfigure[B-SNN Resistor State]{
    \begin{minipage}{1.5in}
    \centering
        \begin{tabular}{||c|c||} 
            \hline
            Weight & R \\ 
            \hline\hline
            -1 &  HRS \\ 
            \hline
            +1 &  LRS \\
            \hline
        \end{tabular}  \label{tab:SNN_res}
    \end{minipage}
}
\caption{BNN vs B-SNN RRAM based ``In-Memory" computing kernel.}
\end{figure}

Considering that the SNN is operated for $N$ time-steps, the network converges to a Binary-Weight Network as $N\rightarrow \infty$. However, for a finite number of timesteps, we can consider the network to be a discretized ANN, where the weights are binary but the neuron activations are represented by $B=log_2{N}$ number of bits. However, since the neuron states are represented by $0$ and $1$ values, B-SNNs are event-driven, thereby resulting in power consumption only when triggered, i.e. on receiving a spike from the previous layer. Hence, while the representative bit-precision can be $\sim 7$ bits for networks simulated over 100 timesteps, the network's computational power does not scale-up corresponding to a multi-bit neuron model. This is explained in Fig. \ref{fig:xnor_circuit}-(e). The left-panel depicts a bit-cell for an ``In-Memory" Resistive Random Access Memory (RRAM) based BNN hardware accelerator \cite{yin2019high}. The RRAM can be programmed to either a high resistive state (HRS) or a low resistive state (LRS). The RRAM states and input conditions for $+1,-1$ are tabulated in Fig. 2 and shows the correspondence to the binary dot-product computation. Note that two rows per input are used due to the differential nature $+1,-1$ of the neuron inputs. Hence, irrespective of the value of the input, one of the rows of the array will be active resulting in power consumption. Fig. \ref{fig:snn_circuit} depicts the same array for the B-SNN scenario. Since, in a B-SNN, the neuron outputs are $0$ and $1$, we can use just one row per bit-cell, thereby reducing the array area by $50\%$. Note that a dummy column will be required for referencing purposes of sense amplifiers interfaced with the array \cite{yin2019high}. Additionally, the neuron circuits interfaced with the array need to accumulate the dot-product evaluation over time. Such an accumulation process can be accomplished using digital accumulators \cite{han2017cross}  or non-volatile memory technologies \cite{wijesinghe2018all,sengupta2016proposal}. Note that energy expended due to this accumulation process is minimal in contrast to the overall crossbar power consumption \cite{ankit2017resparc}. 
However, the input to the next layer will be a binary spike, thereby enabling us to utilize the  ``In-Memory" computing block as the core hardware primitive. It is worth noting here that the power-consumption involved in accessing the rows of the array occurs only on a spike event, thereby resulting in event-driven operation. 

\section{Experiments and Results}
\subsection{Datasets and Implementation} 
We evaluate our proposal on two popular, publicly available datasets, namely the CIFAR-$100$ \cite{cifar100} and large-scale ImageNet \cite{deng2009imagenet} dataset. CIFAR-$100$ dataset contains $100$ classes with $60,000$ $32 \times 32$ colored images where $50,000$ images were used for training and $10,000$ images were used for testing. The more challenging ImageNet $2012$ dataset contains $1000$ classes of images of various objects. The dataset contains $1.28$ million training images and $50,000$ validation images. Randomly cropped $224\times 224$ pixel regions were used for the ImageNet dataset. All empirical analysis and optimizations were performed on the CIFAR-$100$ dataset and the resultant conclusions and settings were used for the final ImageNet simulation. All experiments are run in PyTorch framework using two GPUs. For both datasets, the image pixels were normalized to have zero mean and unit variance. Other standard pre-processing techniques used in this work can be found at \href{https://github.com/NeuroCompLab-psu/SNN-Conversion}{(Link)}. It is worth mentioning here that while evaluations in this paper are based on static datasets, SNNs are inherently suited for spatio-temporal datasets generated from event-driven sensors \cite{li2017cifar10,amir2017low} and such sensor-algorithm co-design is currently an active research area \cite{ruckauer2019closing}.

Our network architecture follows a standard VGG-$16$ model. We purposefully chose the VGG architecture since many of the inefficiencies and accuracy degradation effects of BNNs are not reflected in shallower models like AlexNet or already-compact models like ResNet. However, we observed that VGG XNOR-Nets could not be trained successfully with $3$ fully connected layers at the end. Hence, to reduce the training complexity, we considered a modified VGG-$15$ structure with one less linear layer. Note that only top-1 accuracies are reported in the paper.

As mentioned earlier, we used ANN-SNN conversion technique to generate our B-SNN. While ANN-SNN conversion is currently the most scalable technique to train SNNs, it suffers from high inference latency. However, recent work has shown SNNs trained directly through backpropagation are characterized by much lower latency than networks obtained through ANN-SNN conversion, albeit for simpler datasets and shallower networks \cite{SNNbackPropLee}. Due to the fact that such training schemes are computationally much more exhaustive, a follow-up work has explored a hybrid training approach comprising ANN-SNN conversion followed by backpropagation-through-time fine-tuning to scale the latency reduction effect to deeper networks \cite{rathi2020enabling}. However, as we show in this work, the full design space of ANN-SNN conversion has not been fully explored. Prior work on ANN-SNN conversion \cite{sengupta2019going} has mainly considered conversion techniques optimizing accuracy, thereby incurring high latency. In this work, we show that there exists extremely simple control knobs (both at design time and at run time) that can be also used to reduce inference latency drastically in ANN-SNN conversion methods without compromising on the accuracy or involving computationally expensive training/fine-tuning approaches. Since our SNN training optimizations are equally valid for full-precision networks, we report accuracies for full-precision models along with their binary counterparts in order to compare against prior art. 

Our ANNs were trained with constraints of no bias and batch-normalization layers in accordance with previous work \cite{sengupta2019going}. A dropout layer was inserted after every ReLU layer (except those followed by a pooling layer) to aid the regularization process in absence of batch-normalization. Our XNOR and B-SNN networks do not binarize the first and last layers as in previous BNN implementations. We apply the pixel intensities directly as input to the spiking networks instead of an artificial Poisson spike train \cite{rueckauer2017conversion}. Once the ANN is trained, it is converted to an iso-architecture SNN by replacing the ReLUs with IF spiking neuron nodes. The SNN weights are normalized by using a randomly sampled subset of images from the training set and recording the maximum ANN activities. Note that normalization based on SNN activities can be used to further reduce the ANN-SNN accuracy gap \cite{sengupta2019going}. The SNN implementation is done using a modified version of the mini-batch processing enabled SNN simulation framework \cite{saunders2019minibatch} in BindsNET \cite{Hazan_2018}, a PyTorch based package. 
 
\subsection{Training B-SNNs} 
\begin{figure}
\centering
    \includegraphics[width=0.32\textwidth]{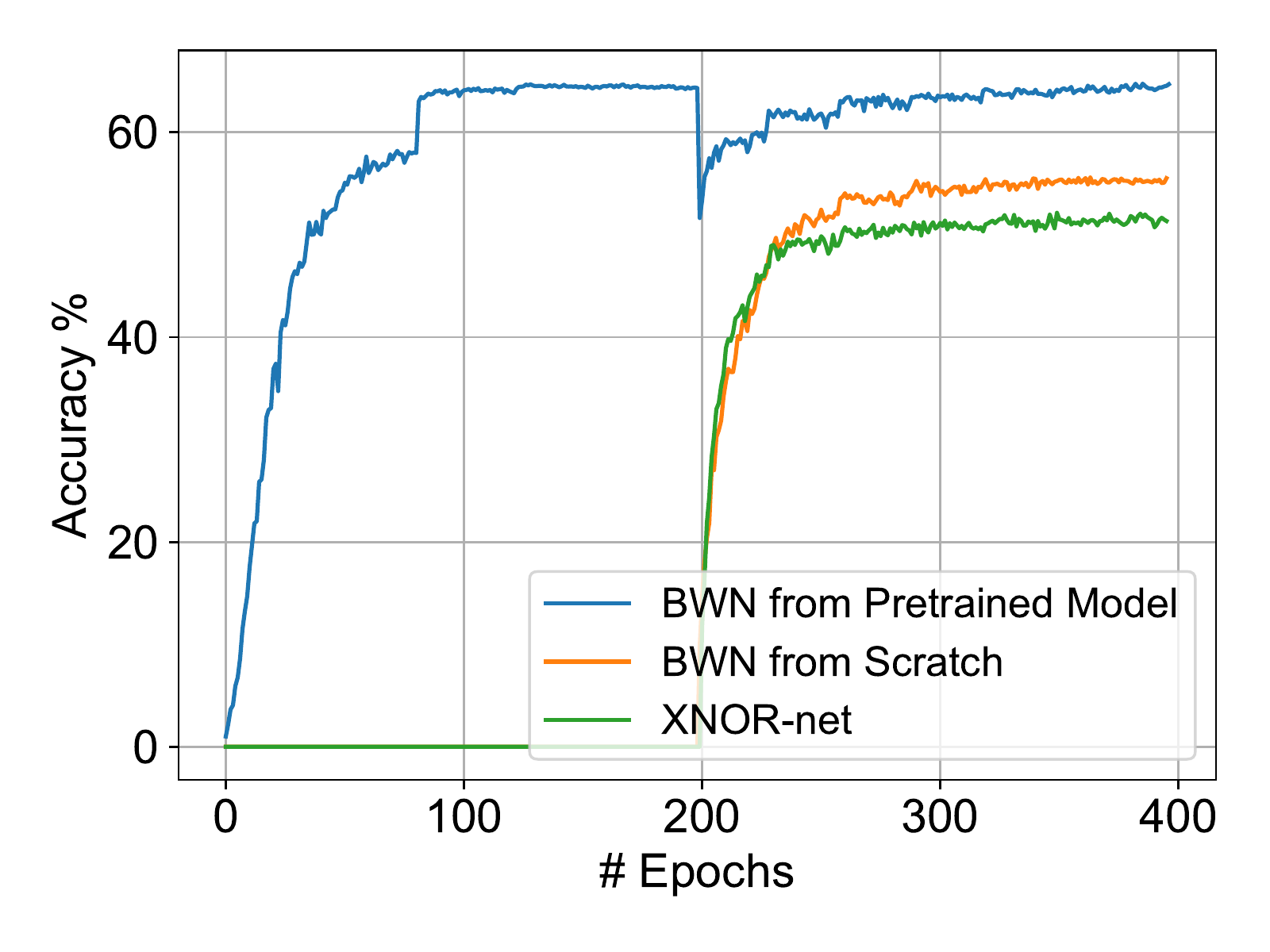}
\caption{Validation results on CIFAR-$100$ dataset. Note that full-precision model training is plotted from $0-200$ epochs. The constrained-BWN model is trained subsequently from the $200$-th epoch. The BWN model trained from scratch and XNOR-Net convergence plots are also shown for comparison.}
\label{fig:avg_train}
\end{figure}

In order to train the B-SNN, we first trained a constrained-BWN, as mentioned previously. ADAM optimizer is used with an initial learning rate of $5e-4$ and a batch size of 128. Lower learning rates for training binary nets have proven to be also effective in a recent study \cite{tang2017train}. The learning rate is subsequently halved every $30$ epochs for a duration of $200$ epochs. The weight decay starts from $5e-4$ and is then set to $0$ after 30 epochs similar to XNOR-Net training implementations \cite{rastegari2016xnor}. As shown in Fig. \ref{fig:avg_train}, we find that the final validation accuracy improvement for the constrained-BWN is minimal over an iso-architecture XNOR-Net. This is primarily due to the constrained nature of models suitable for ANN-SNN conversion coupled with weight binarization.

However, previous work has indicated that careful weight initialization is crucial for training networks without batch-normalization \cite{sengupta2019going}. Drawing inspiration from that observation, we performed a hybrid training approach, where a constrained full-precision model was first trained and then subsequently binarized with respect to the weights. The resultant constrained-BWNs exhibited accuracies close to original full-precision accuracies, as shown in Fig. \ref{fig:avg_train}. A similar hybrid training approach was also recently observed to speed up the training process for normal BNNs \cite{alizadeh2018empirical}. Note that the full precision networks are trained for $200$ epochs with a batch size of $256$, an initial learning rate of $5e-2$, weight decay of $1e-4$ and SGD optimizer with a momentum of $0.9$. The learning rate was divided by $10$ at $81$ and $122$ epochs. The trained full-precision models are also used for substantiating the benefits of the SNN optimization control knobs discussed next.

\subsection{Design-Time SNN Optimizations}

\subsubsection{Architectural Options} An important design option in the SNN/BNN architecture is the type and location of pooling mechanism. Normal deep networks usually have pooling layers after the neural node layer to compress the feature map.  Among the two options typically used - Max Pooling and Average Pooling - architectures with Max Pooling are usually characterized by higher accuracy. However, because of the binary nature of neuron outputs in BNN/SNN, Max Pooling after the neuron layer should result in accuracy degradation. To circumvent this issue, BNN literature has explored using Max Pooling before the neuron layer \cite{rastegari2016xnor} while SNN literature has considered Average Pooling after the neuron layer \cite{sengupta2019going}. A comprehensive analysis in this regard is missing. 

\begin{figure}[htp]
  \centering
  \subfigure[]{\includegraphics[scale=0.25]{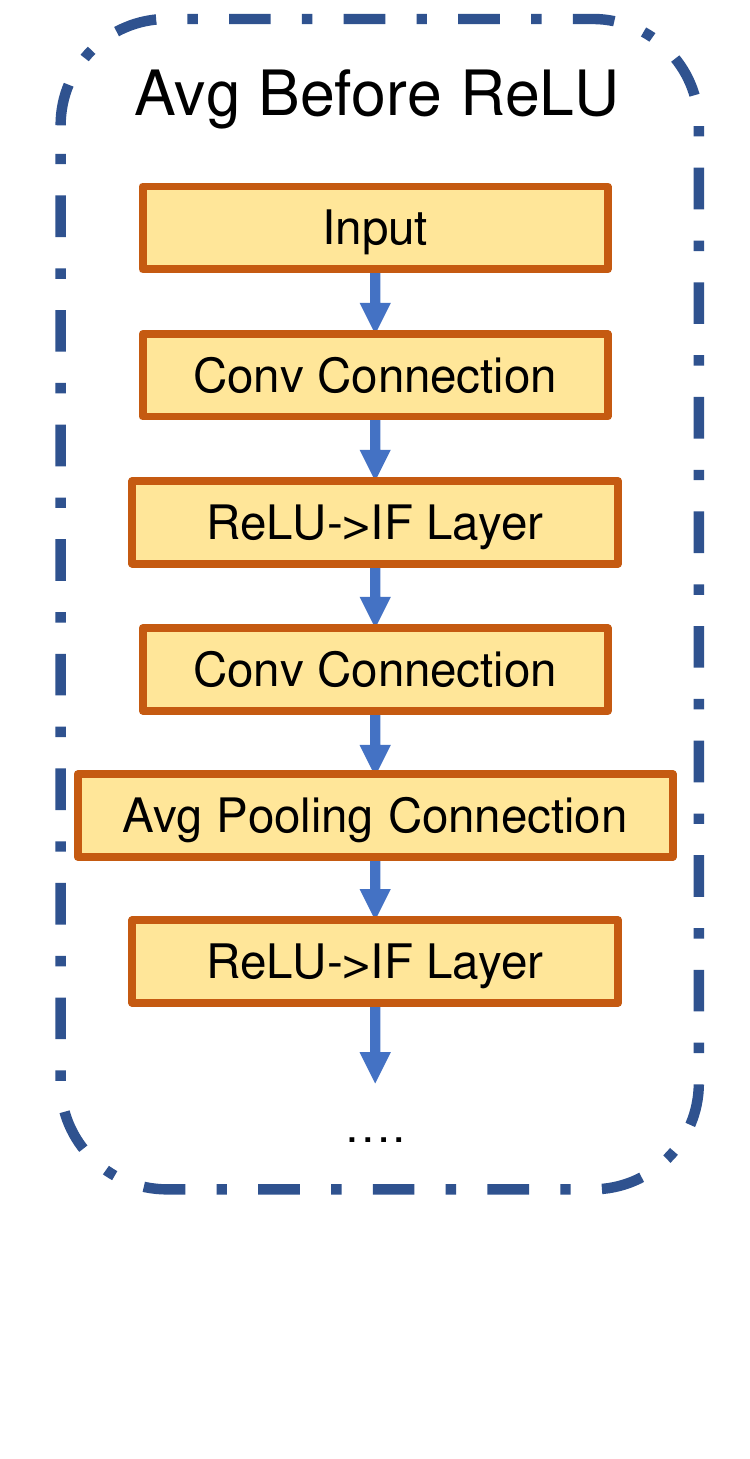}\label{fig:ab_struc}}\quad
  \subfigure[]{\includegraphics[scale=0.25]{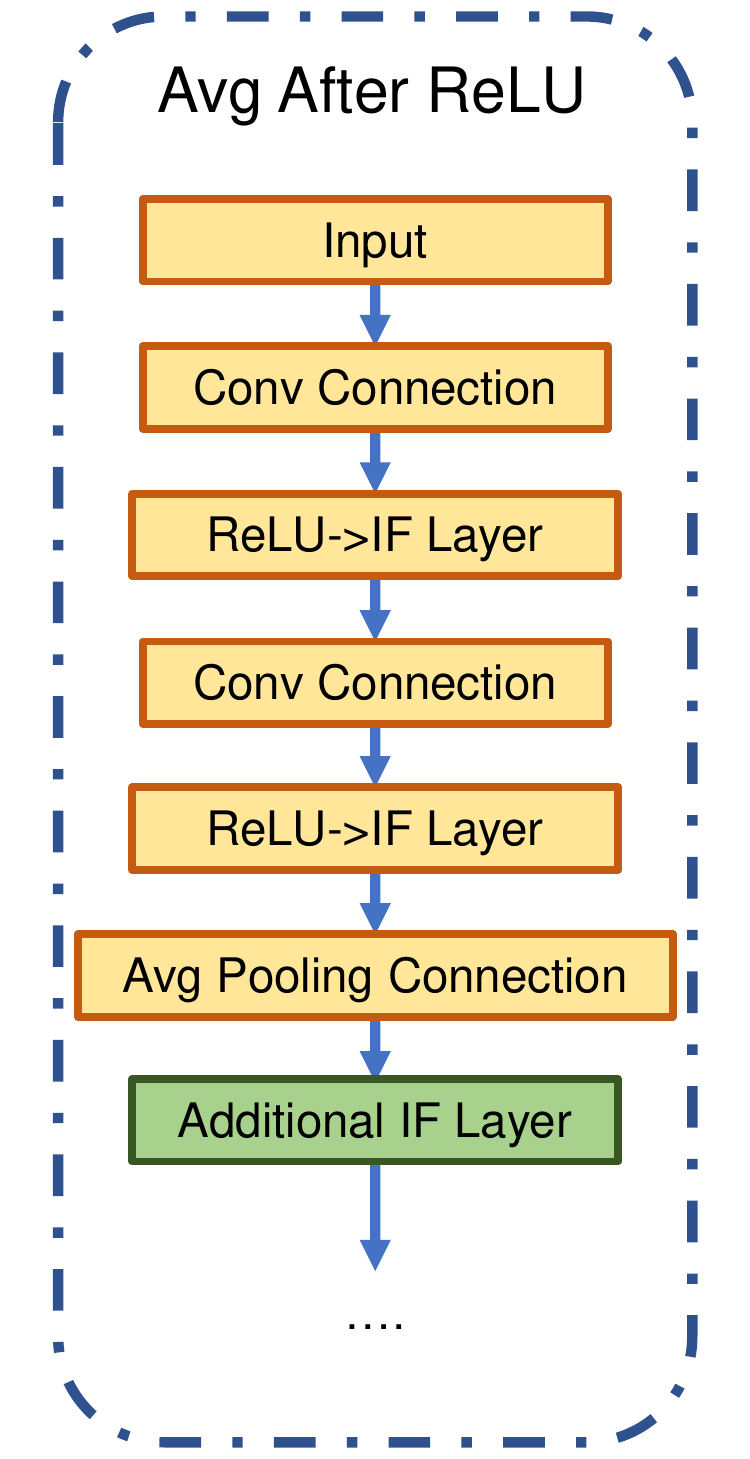}\label{fig:aa_struc}}
  \subfigure[]{\includegraphics[scale=0.25]{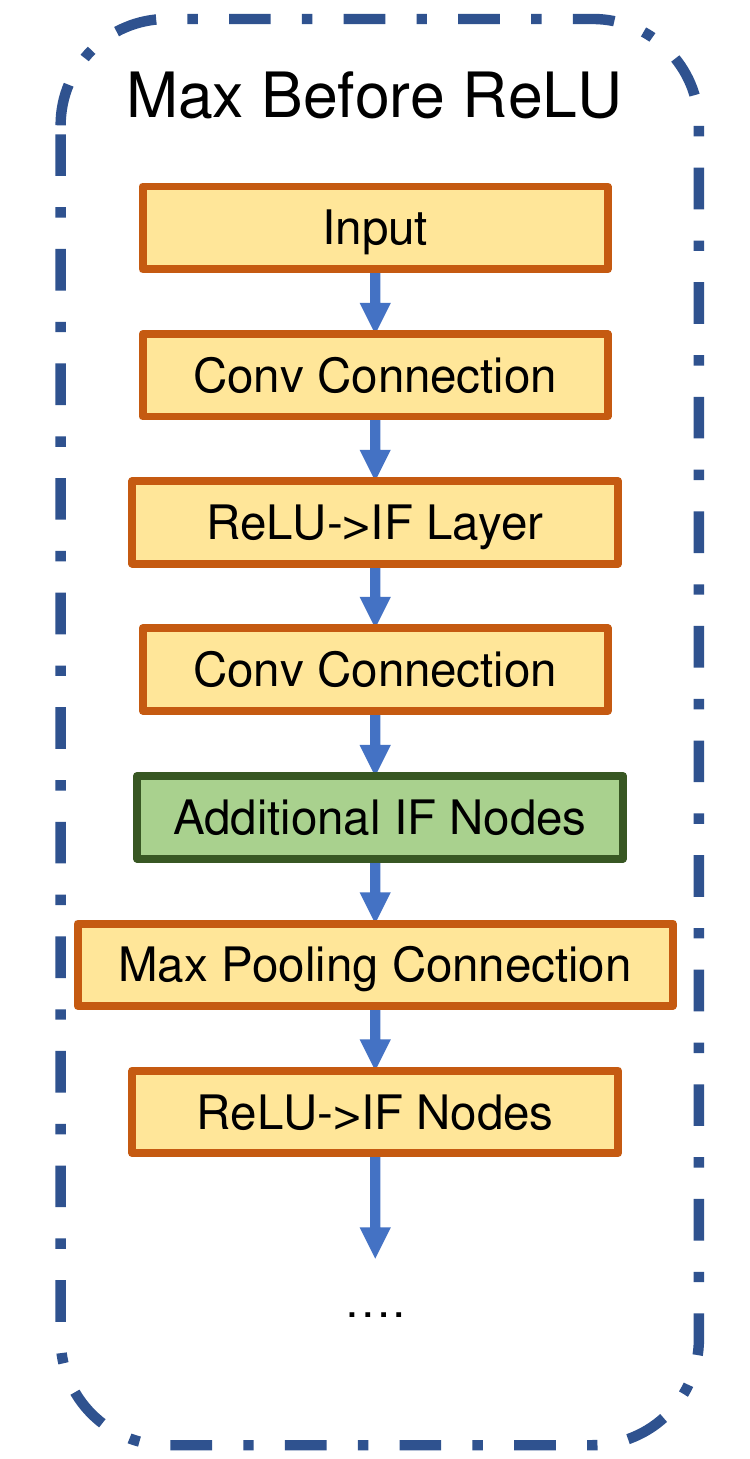}\label{fig:mb_struc}}\quad
  \subfigure[]{\includegraphics[scale=0.25]{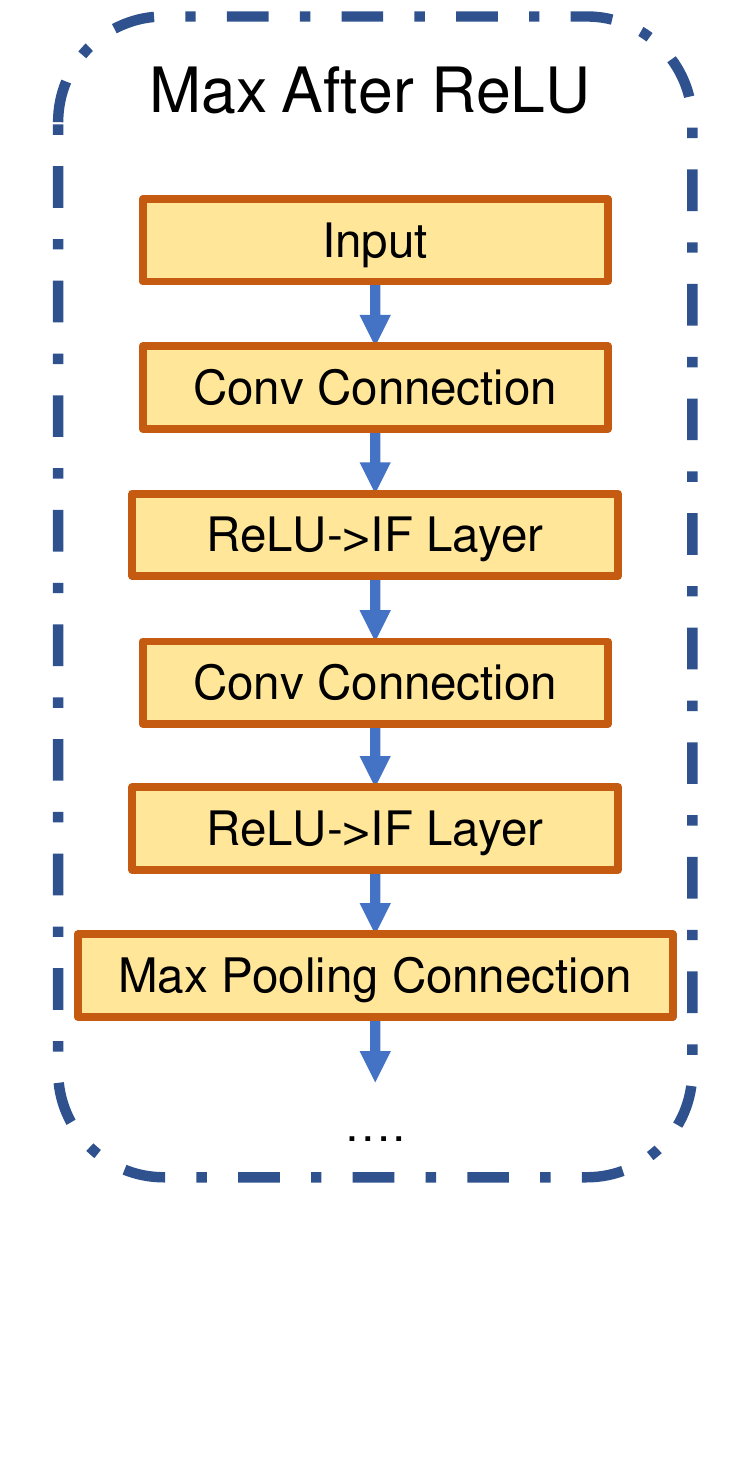}\label{fig:ma_struc}}
  \caption{Network Architectural Options: Average/Max-Pooling before/after ReLU/IF neuron layers.}
  \label{fig:a2s_conversion}
\end{figure}

In this work, we trained network architectures with four possible options - Average/Max-Pooling before/after the ReLU/IF neuron layer (Fig. \ref{fig:a2s_conversion}). All four constrained-BWN architectures perform similarly on CIFAR-$100$, as full-precision ANNs, and converge to accuracy of $64.9\%, 65.8\%, 67.7\%$ and $67.6\%$ for Average-Pooling before and after ReLU, Max-Pooling before and after ReLU respectively. As expected, the Max-Pooling architectures perform slightly better. However, converted SNNs with Max-Pooling would result in accuracy degradation during the conversion process since the max-pooling operation is not distributed linearly over time. In contrast, the linear Avg-Pooling operation would not involve such issues during the conversion process. This tradeoff was evaluated in this design space analysis. We would like to mention here that two architectural modifications were performed while converting the constrained-BWN to B-SNN. First, as shown in Fig. \ref{fig:aa_struc}, an additional IF layer was added after the Average-Pooling layer to ensure that the input to the next Convolutional layer is binary (to utilize the underlying binary hardware primitive). Also, for the Max-Pooling before ReLU option (Fig. \ref{fig:mb_struc}), we inserted an additional IF neuron layer after the Convolutional layer. We observed that absence of this additional layer resulted in extremely low SNN accuracy ($33\%$). We hypothesize this to occur due to Max-Pooling the Convolutional outputs directly over time at every time-step.

\begin{figure}[htp]
  \centering
  \subfigure[Full-Precision SNN]{\includegraphics[scale=0.26]{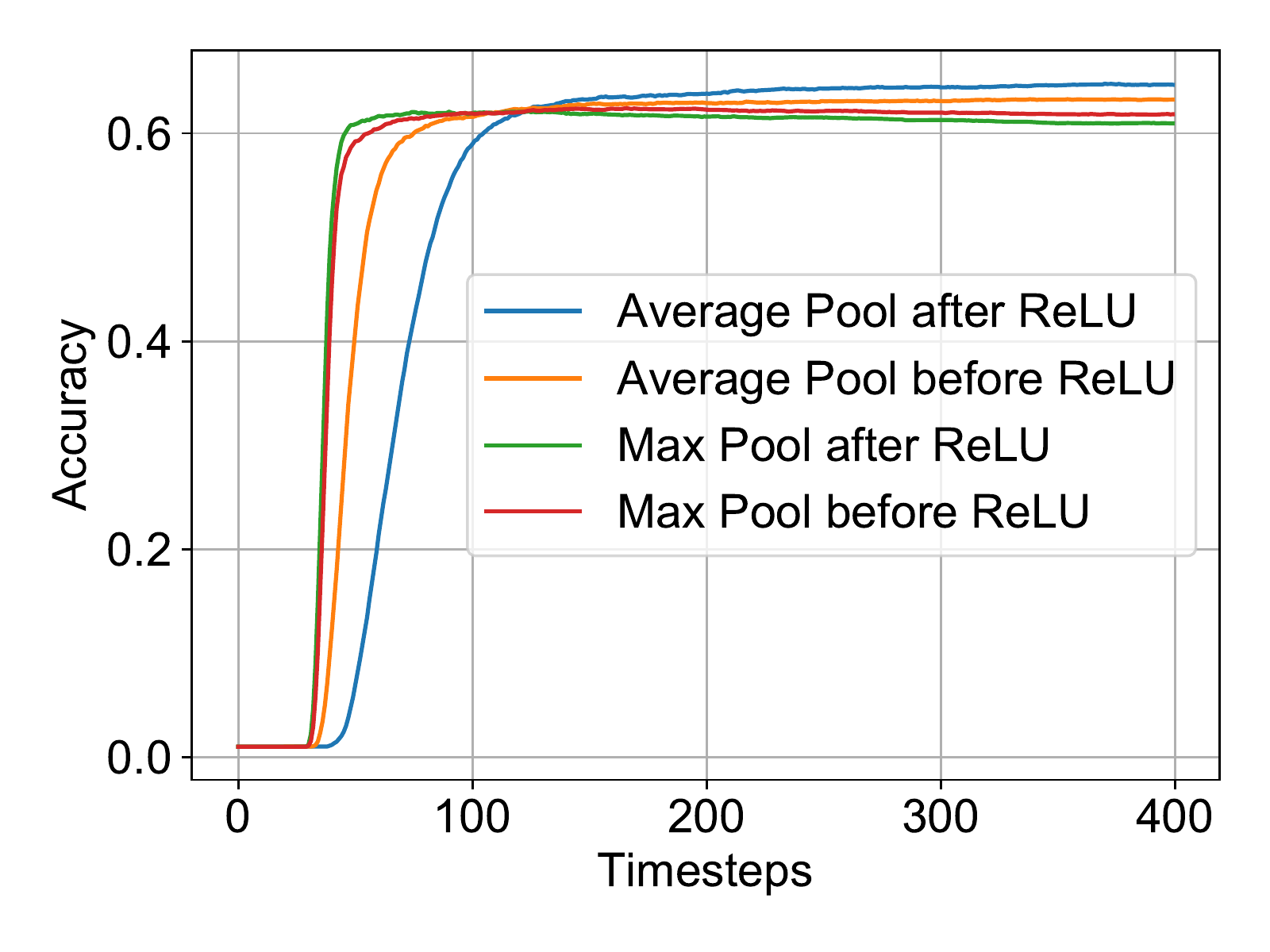}\label{fig:fp_snn}}\quad
  \subfigure[Binary SNN]{\includegraphics[scale=0.26]{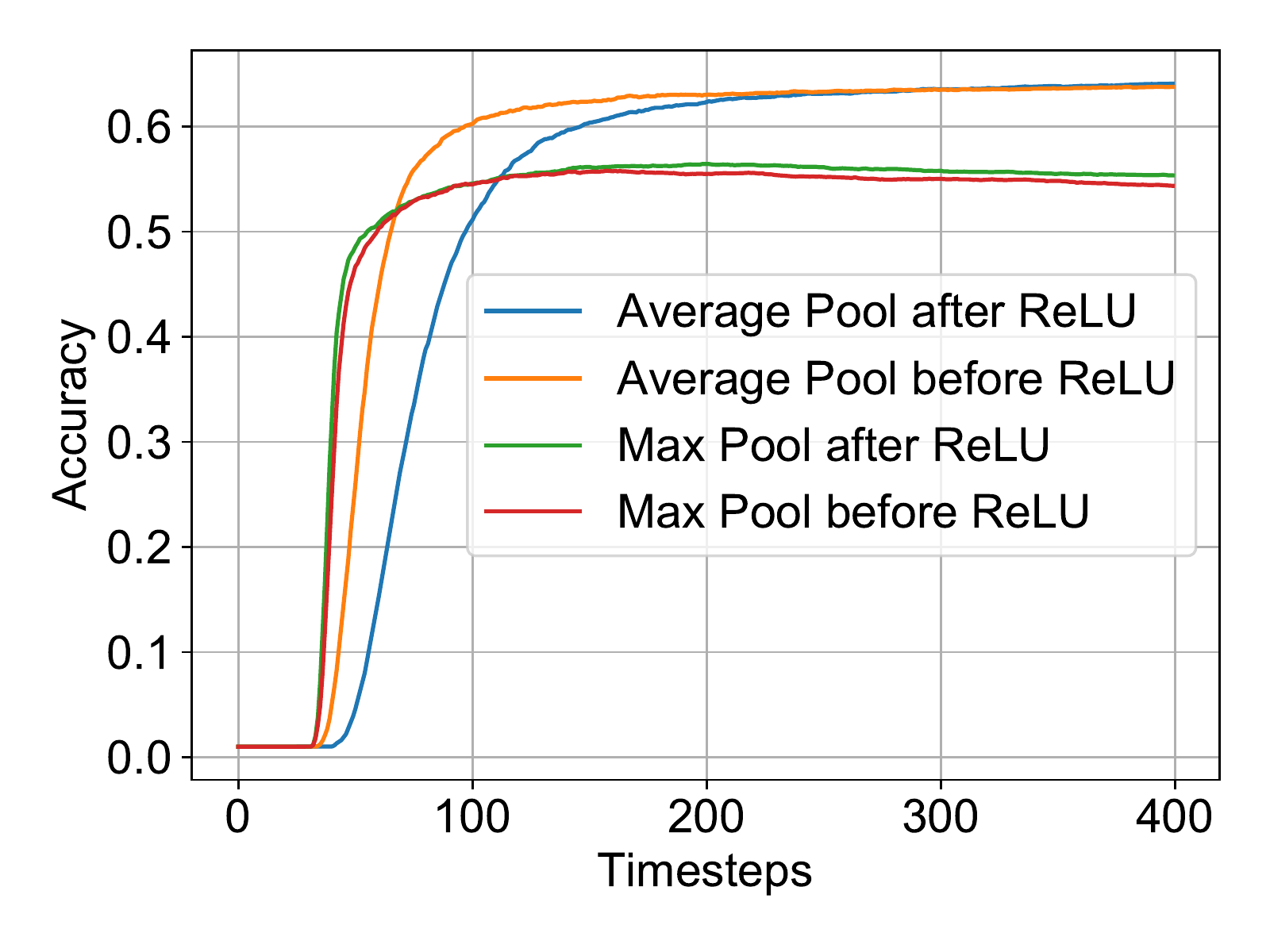}\label{fig:b_snn}}
  \caption{SNN Accuracy variation with time-steps on CIFAR-$100$ dataset for various architectural options. Note that all SNNs used here have Subtractive-IF layers.}
  \label{fig:all_avt}
\end{figure}

The variation of SNN accuracy with time-steps is plotted in Fig. \ref{fig:all_avt} for full-precision and B-SNN models respectively. While the baseline ANN Max-Pooling architectures provide better accuracies, they undergo higher accuracy degradation during the conversion process. For the Average-Pooling models, the option with pooling after the neuron layer have higher latency due to additional spiking neuron layers. We find that the Average-Pooling before ReLU/IF neuron layer offers the best tradeoff between inference latency and final accuracy. We therefore chose this design option for the next set of experiments. Note that Fig. \ref{fig:avg_train} shows the convergence graph for this architecture. Similar variation was also observed for the other options. For this architecture option, the full-precision (binary) SNN accuracy is \textbf{63.2\%} (\textbf{63.7\%}) in contrast to full-precision (binary) ANN accuracies of \textbf{64.9\%} (\textbf{64.8\%}).

\subsubsection{Neural Node Options} Another underexplored SNN architecture option is the choice of the spiking neuron node. While prior literature has mainly considered IF neurons where the membrane potential is reset upon spiking, Ref. \cite{rueckauer2017conversion} considers the membrane potential subtracted by the threshold voltage at a firing event. We will refer to the two neuron types as Reset-IF (RIF) and Subtractive-IF (SIF) respectively. SIF neurons assist in reducing the accuracy degradation of converted SNNs by removing the discontinuity occurring in the neuron function at a firing event \cite{rueckauer2017conversion}. However, this is achieved at the cost of higher spiking activity. We would like to stress here that while SNNs reduce the power consumption due to time-domain redistribution of computation, optimizing the SNN energy consumption is a tradeoff between the power benefits and latency overhead -  which is a function of such architectural options considered herein. 

For our analysis, we consider the following proxy metrics for the energy consumption of the ANN and SNN. Assuming that the major energy consumption would occur in the ``In-Memory" crossbar arrays discussed previously, the energy consumption of the ANN will be proportional to the sum of the number of operations in the convolutional and linear layers (due to corresponding activations of the rows of the crossbar array). However, in case of SNNs, the operation is conditional in the case of a spiking event. The calculation for ANN operations in convolutional and linear layers are performed using Eqs. \ref{eq:conv_calculation}-\ref{eq:linear_calculation}. 
\begin{equation}
\mbox{Convolution Layer \#OPS} = nIP * kH * kW * nOP * oH * oW
\label{eq:conv_calculation}
\end{equation}
\begin{equation}
\mbox{Linear Layer \#OPS} = iS*oS
\label{eq:linear_calculation}
\end{equation}
where, $nIP$ is the number of input planes, $kH$ and $kW$ are the kernel height and width, $nOP$ is the number of output planes, $oH$ and $oW$ are the output height and width, and $iS$ and $oS$ are the input and output sizes for linear layers.

In order to measure the efficiency of the SNN with respect to ANN in terms of energy consumption, we use the following Normalized Operations count henceforth.
\begin{equation}
\mbox{Normalized \#OPS} = \frac{\sum_{i=2}^{L-1} \mbox{IFR}_{i} * \mbox{Layer \#OPS}_{i+1}}{\sum {\mbox{Layer \#OPS}}}
\label{eq:normalized_ops}
\end{equation}
where, IFR stands for IF Spiking Rate (total number of spikes over the inference time window averaged over number of neurons), and Layer \#OPS include the operation counts in convolution and linear layers. $L$ represents  the total number of layers in the network. Note that, lower the value of normalized operations, higher is the energy efficiency of converted SNN, with 1 reflecting iso-energy case. Note that we do not consider the operation count for the first and last layers since they are not binarized.

\begin{figure}[htp]
  \centering
  \subfigure[Accuracy versus timesteps for full-precision model.]{\includegraphics[scale=0.26]{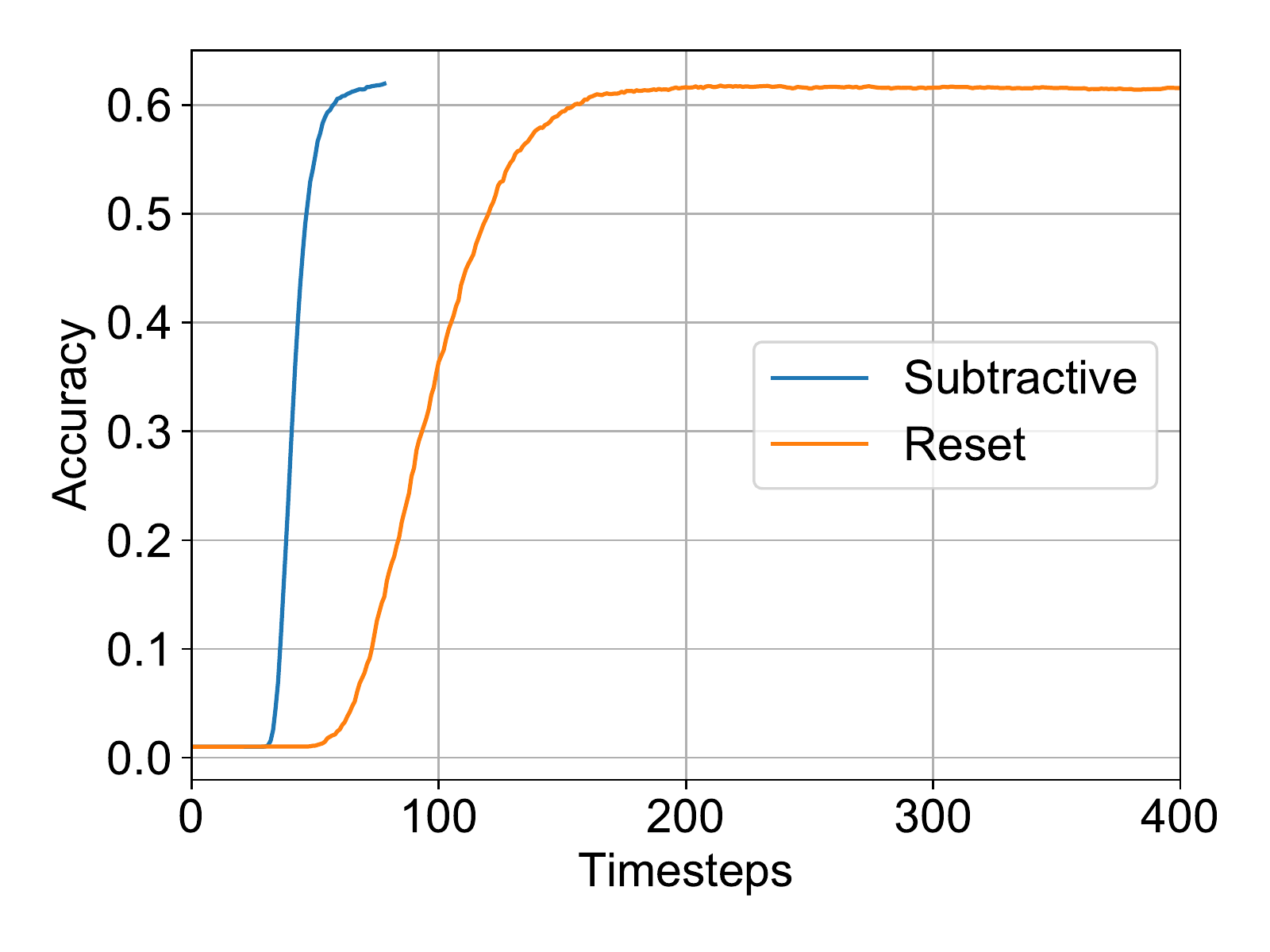}\label{fig:reset_acc_fp}}\quad
  \subfigure[Layerwise IFR for full-precision model.]{\includegraphics[scale=0.26]{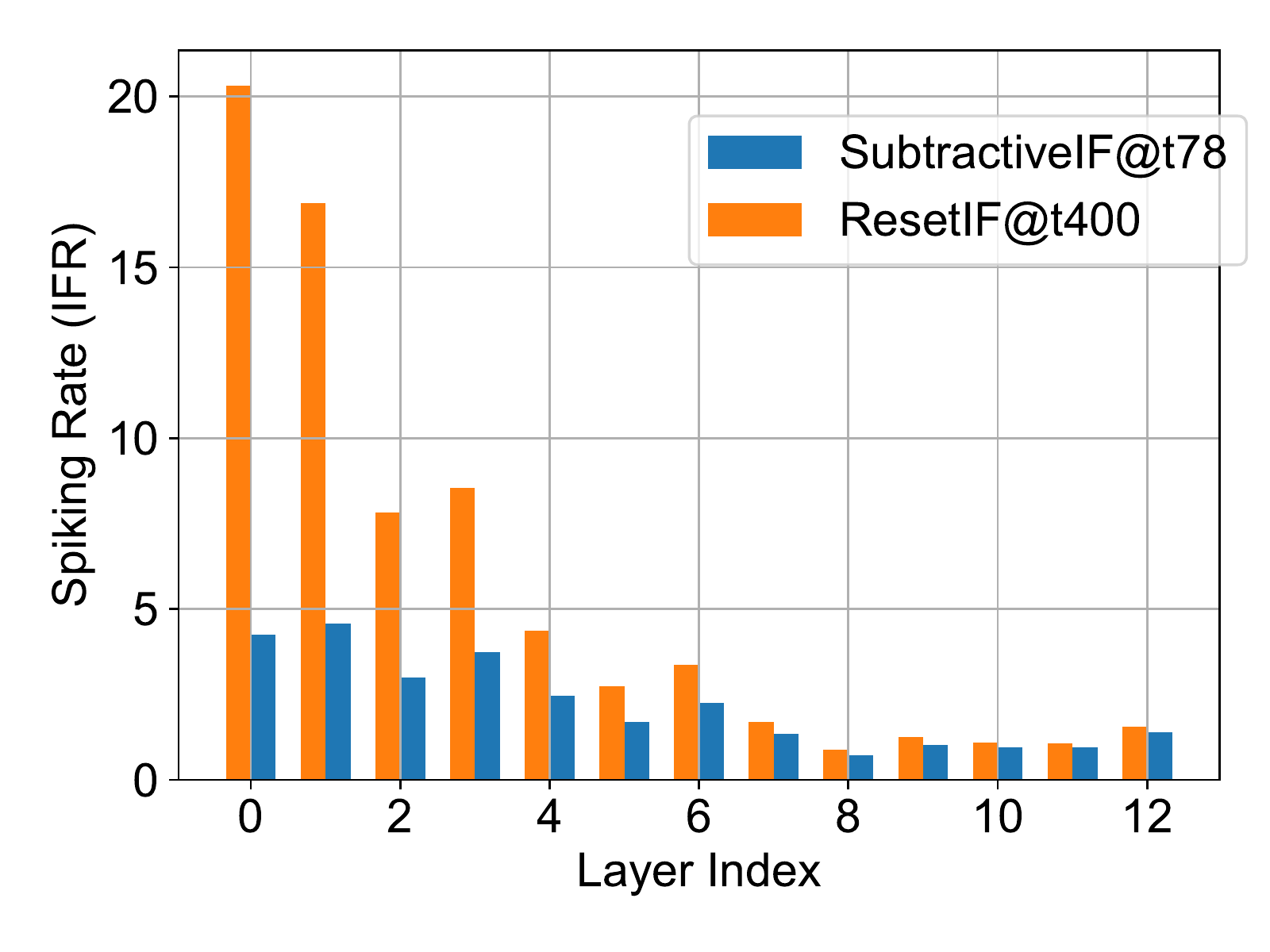}\label{fig:spikes_fp}}
  \label{fig:fp_reset}
  \subfigure[Accuracy versus timesteps for binary model.]{\includegraphics[scale=0.26]{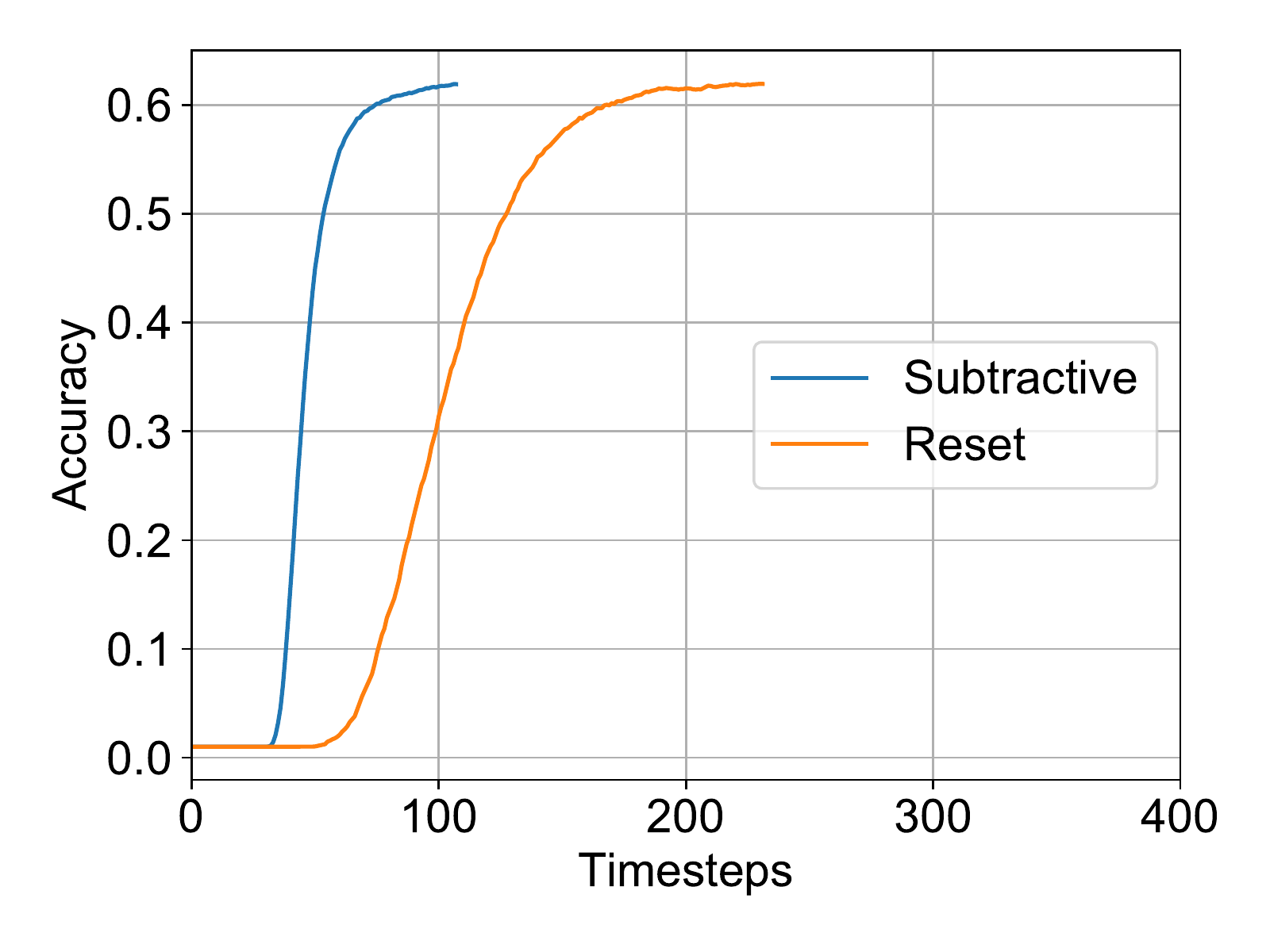}\label{fig:reset_acc_bsnn}}\quad
  \subfigure[Layerwise IFR for binary model.]{\includegraphics[scale=0.26]{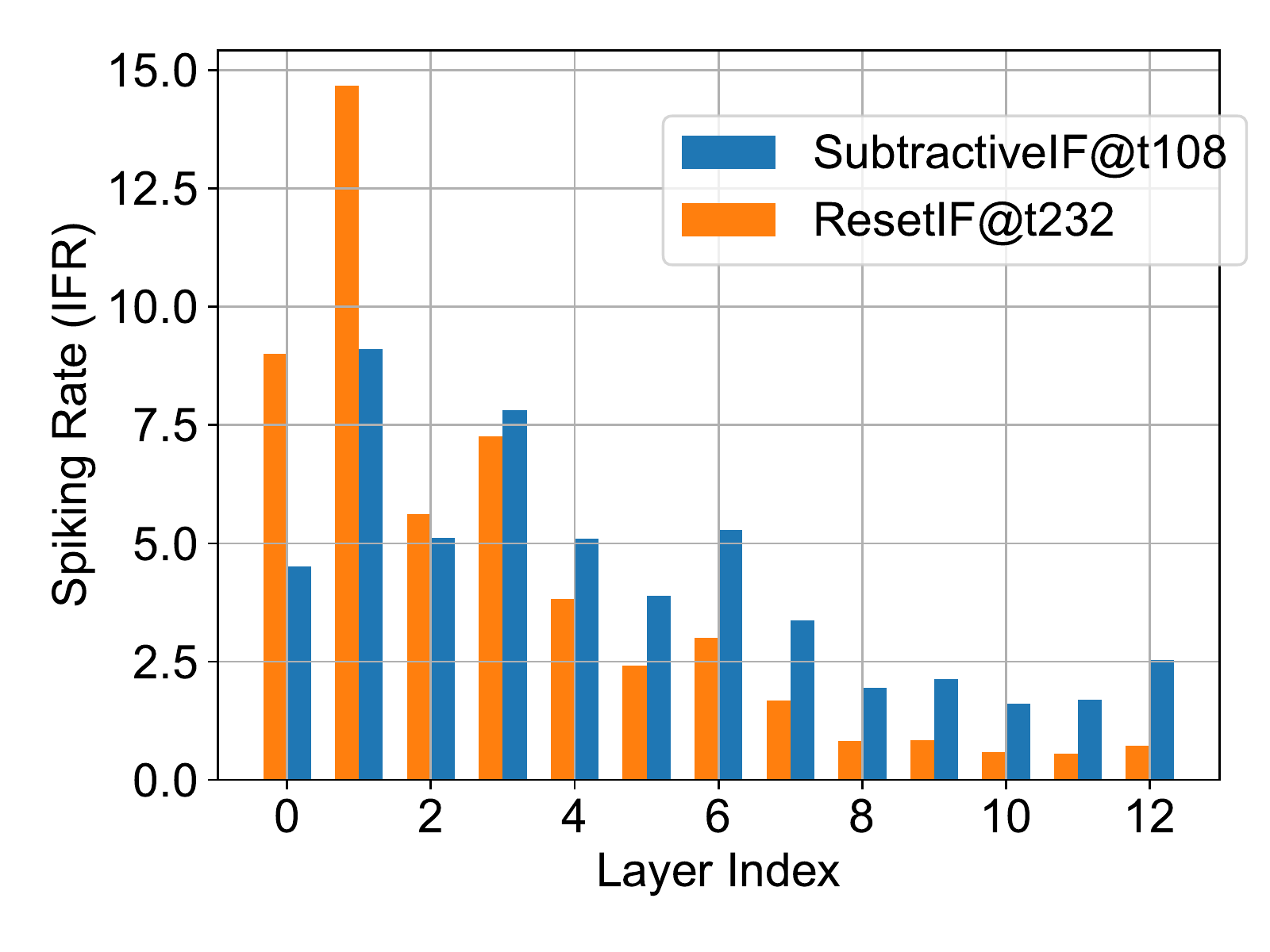}\label{fig:spikes_bsnn}}
  \caption{Analysis for neural node options - SIF versus RIF.}
  \label{fig:bsnn_reset}
\end{figure}

Considering a baseline accuracy of $62\%$, Figs. \ref{fig:bsnn_reset}(a) and 6(c) shows that the SNN structure with SIF has a much smaller delay than the RIF structure. This is intuitive since the spiking rate is much higher in SIF due to subtractive reset. We also observed that the RIF topology was more error-prone during conversion due to the discontinuity occurring on reset to zero. For instance, the full-precision RIF SNN model was unable to reach $62\%$ during 400 timesteps. The total number of normalized operations for SIF and RIF are 4.40 and 4.38 respectively for the B-SNN implementation, and 2.35 and 6.40 (did not reach $62\%$ accuracy) respectively for the full-precision SNN. The layerwise spiking activity is plotted in Figs. \ref{fig:spikes_fp} and Fig. \ref{fig:spikes_bsnn} (the numbers in figure inset represent the timesteps required to reach $62\%$ accuracy). Since the number of computations does not greatly increase for the SIF model with significantly less delay and better accuracy, we choose the SIF model for the remaining analysis.

\subsection{Run-Time SNN Optimizations}

\subsubsection{Threshold Balancing Factor} 
\begin{figure}[htp]
  \centering
  \subfigure[Histogram]{\includegraphics[scale=0.155]{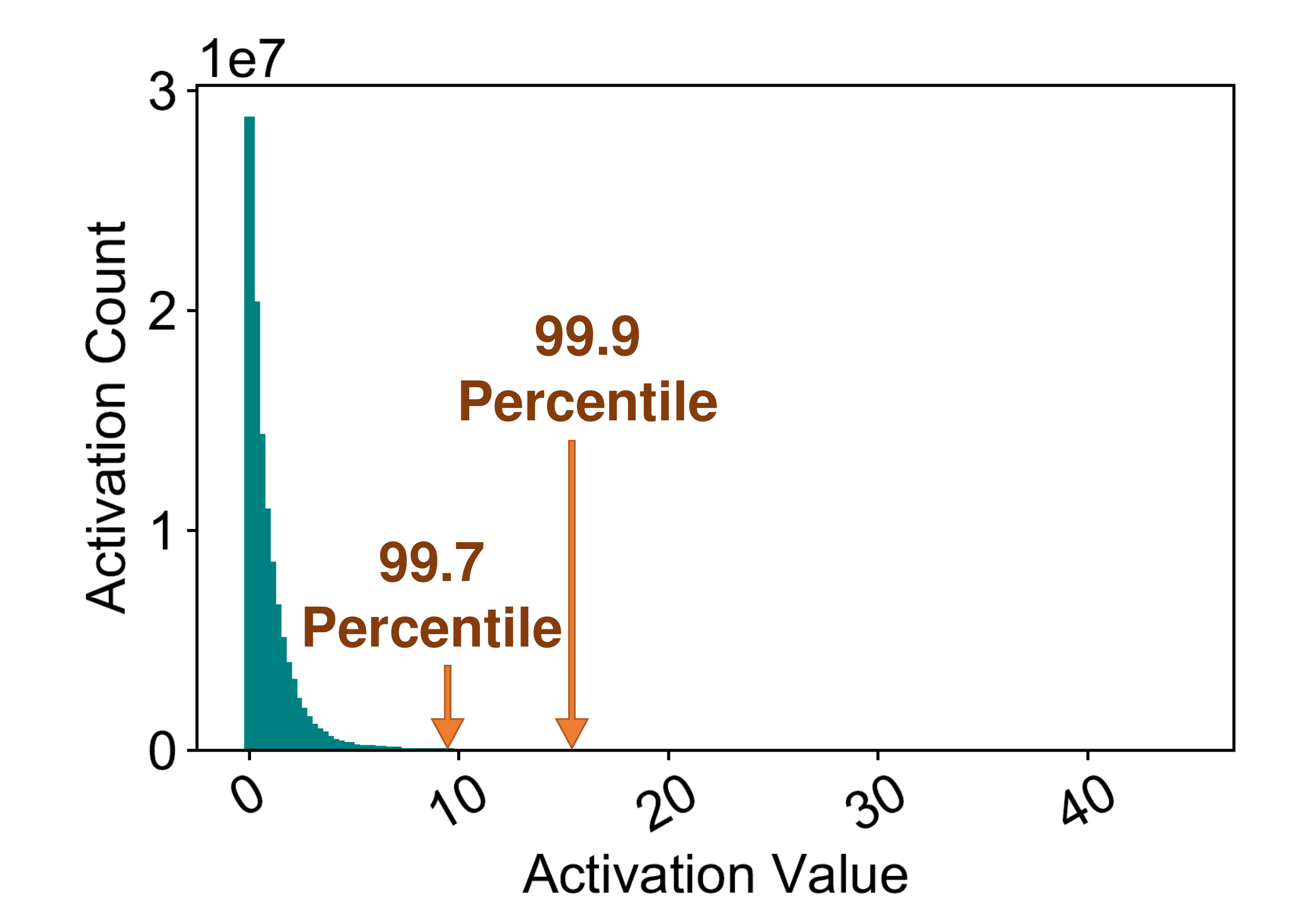}\label{fig:act}}\quad
  \subfigure[Histogram (a) in log-scale]{\includegraphics[scale=0.155]{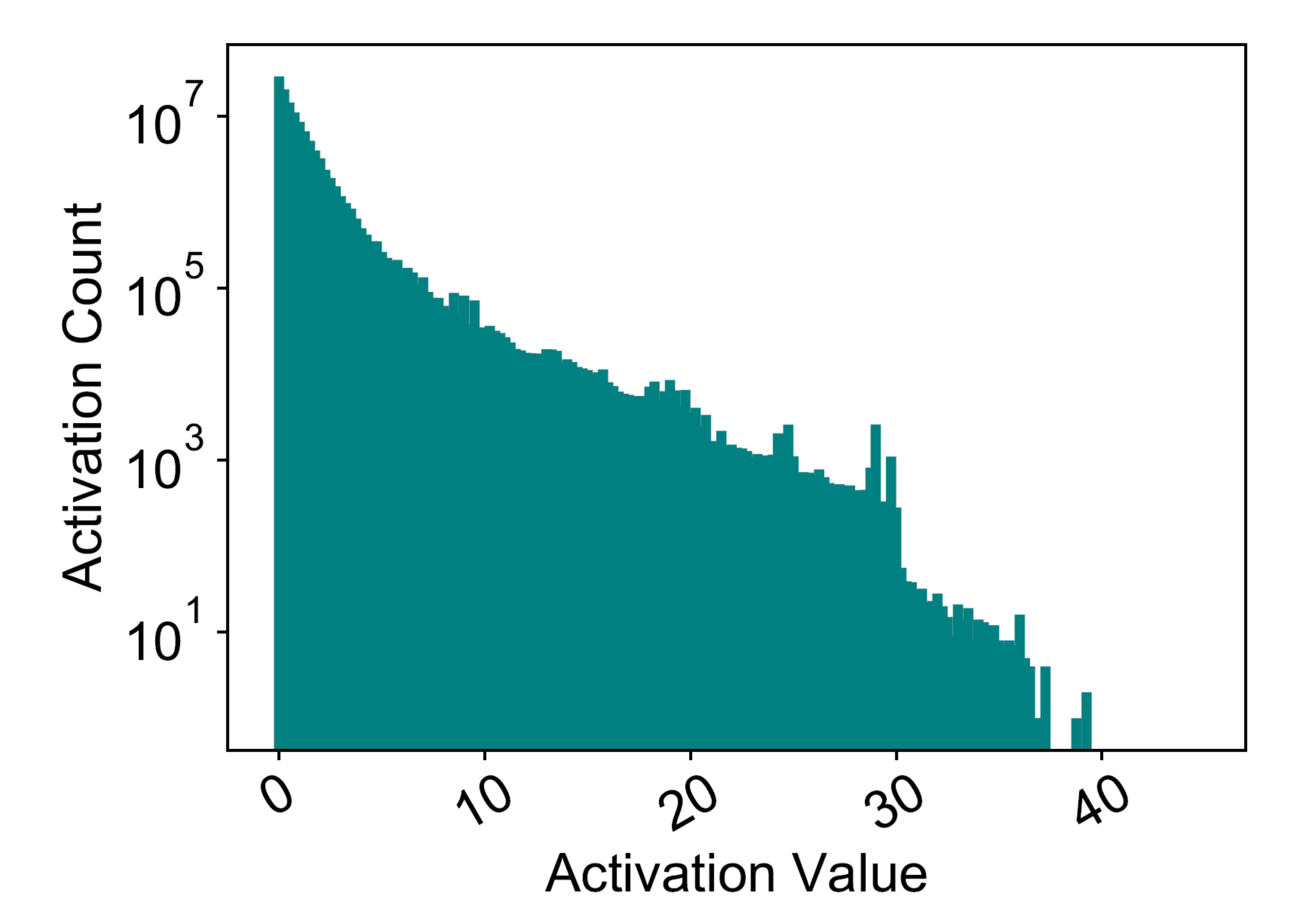}\label{fig:act_log}}
  \caption{Maximum ANN activations for a particular layer.}
  
\end{figure}

Prior work has usually considered the maximum activation of the ANN/SNN neuron as the neuron firing threshold for a particular layer, as explained in Section IV-C. Fig. \ref{fig:act} plots the histogram of the maximum ANN activations of a particular layer. The distribution is characterized by a long tail (Fig. \ref{fig:act_log}) which results in an unnecessarily high SNN threshold, since most of the actual SNN activations would be much lower at inference time. This observation was consistently observed for other layers. Hence, while prior work has shown ANN-SNN conversion to be characterized by extremely high latency, it is due to the fact that the model is optimized for high accuracy, which translates to high latency. In this work, we analyze the effect of varying the threshold balancing factor by choosing a particular percentile from the activation histogram.

\begin{figure}[htp]
  \centering
  \subfigure[Accuracy versus timesteps for full-precision model.]{\includegraphics[scale=0.26]{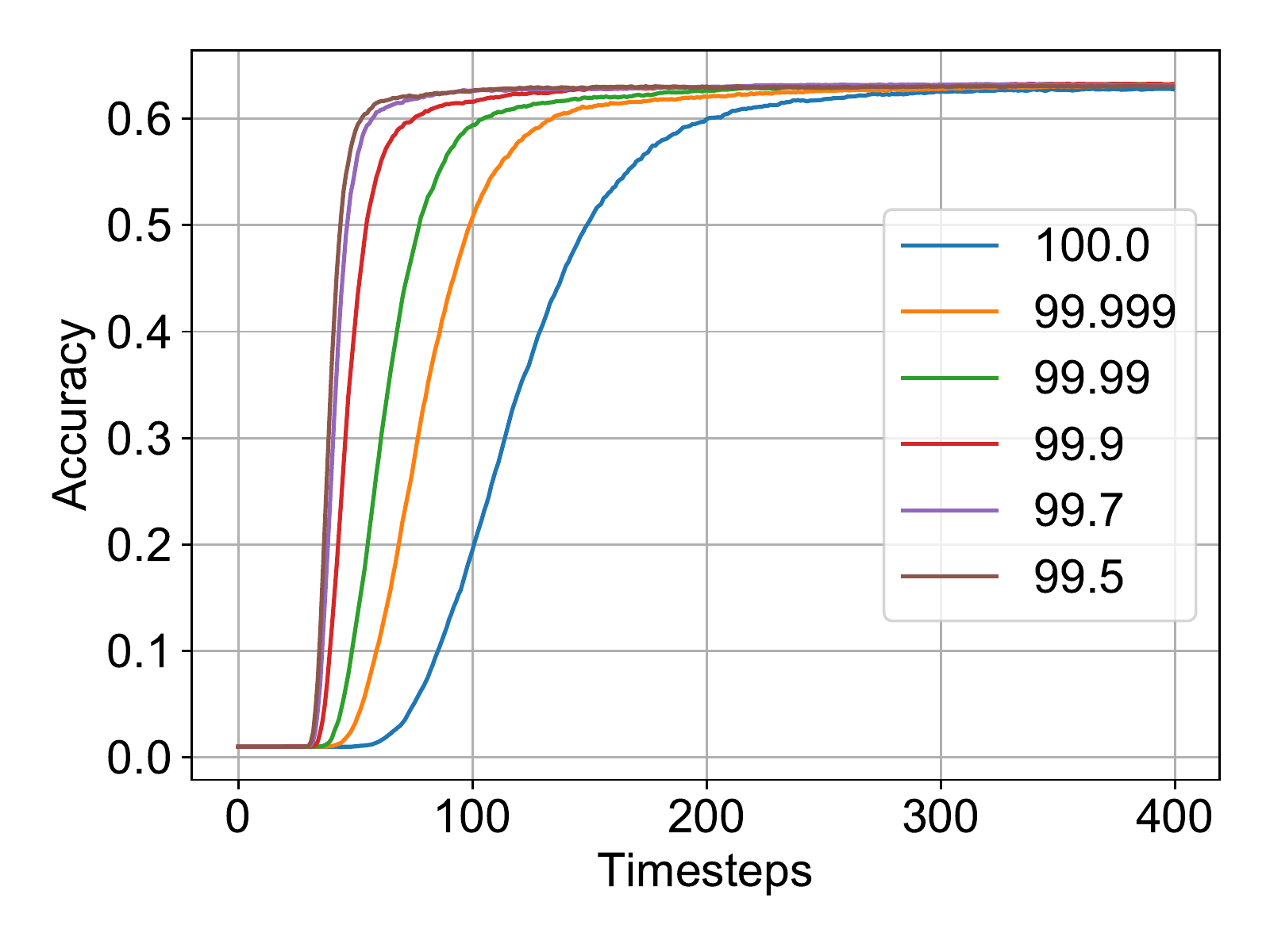}\label{fig:fp_percentile}}\quad
  \subfigure[Normalized \#OPS with varying percentile for the full-precision model. ]{\includegraphics[scale=0.26]{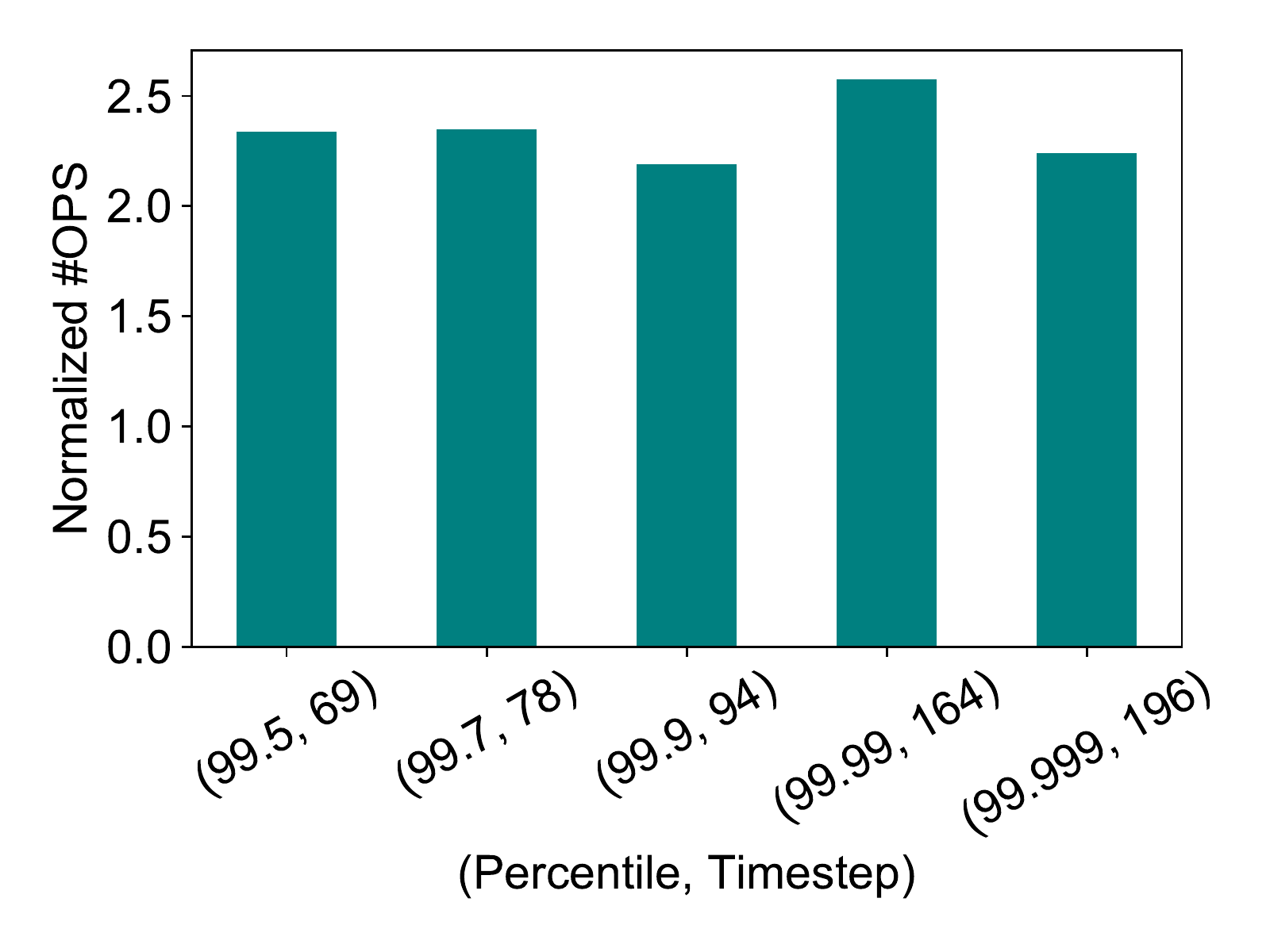}\label{fig:fp_perc_cmp}}
  \subfigure[Accuracy versus timesteps for binary model.]{\includegraphics[scale=0.26]{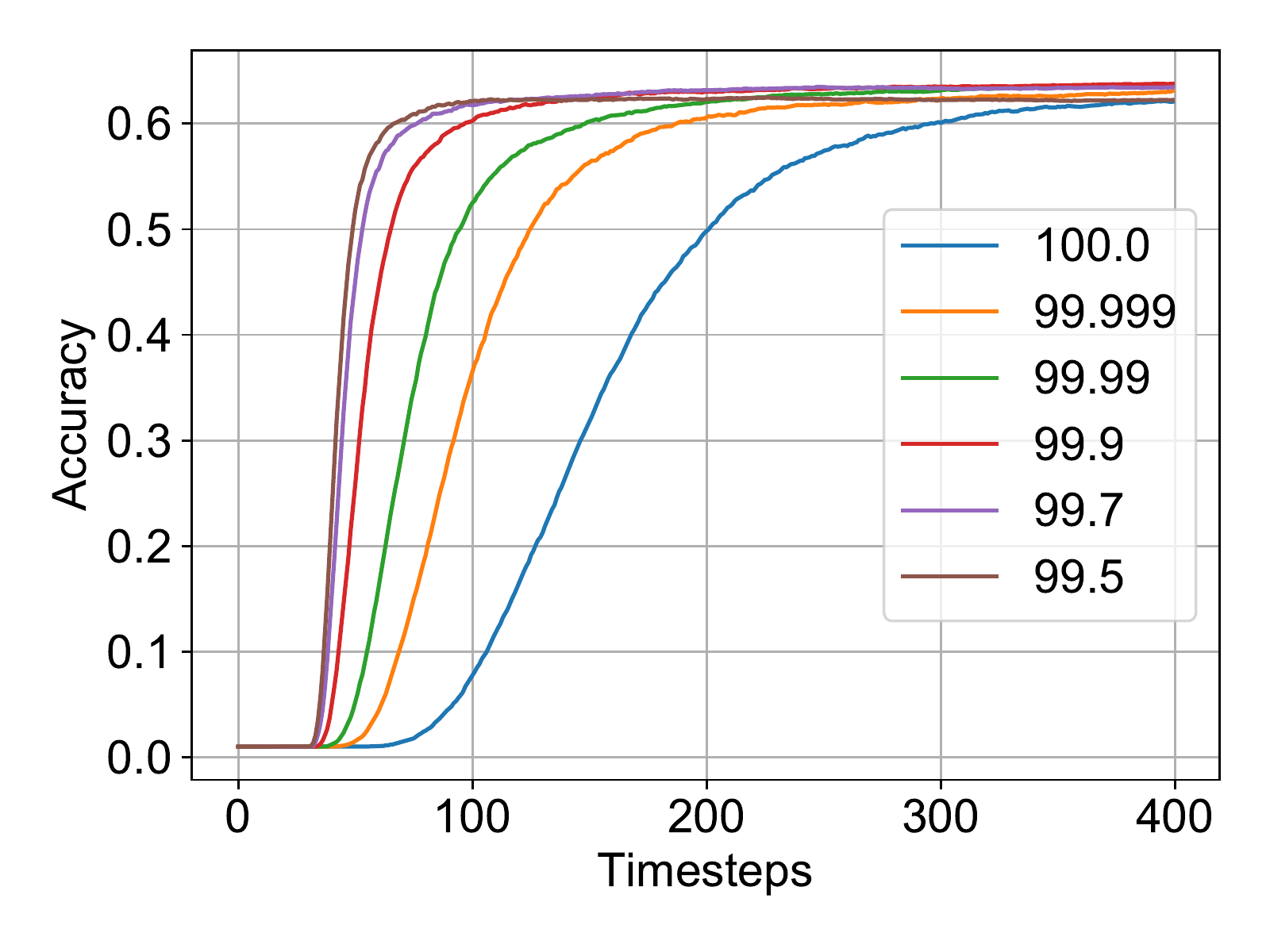}\label{fig:bsnn_percentile}}\quad
  \subfigure[Normalized \#OPS with varying percentile for the binary model.]{\includegraphics[scale=0.26]{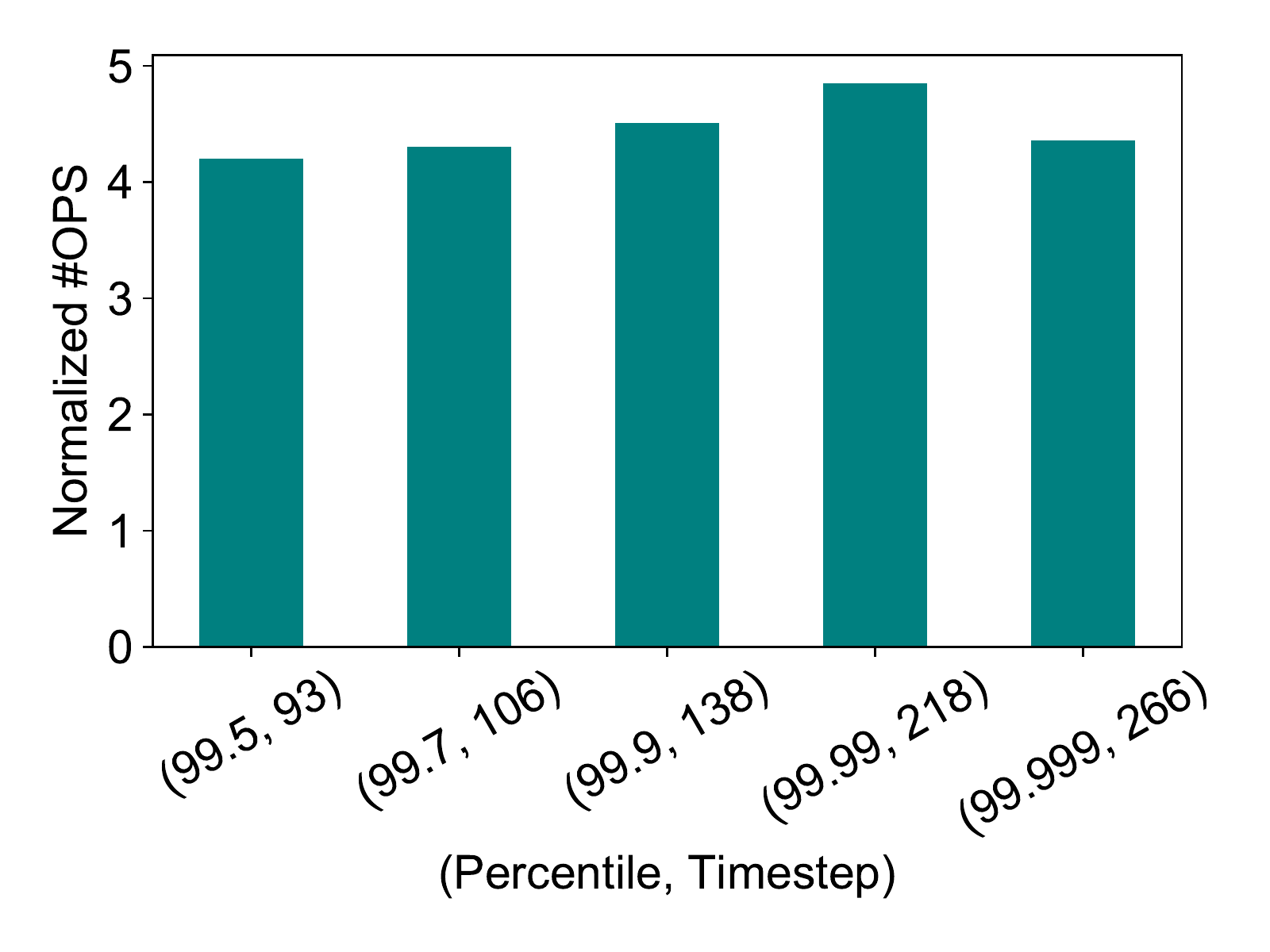}\label{fig:bsnn_perc_cmp}}
  \caption{Analysis for Threshold Balancing Factor.}
  \label{fig:snn_percentile}
\end{figure}

Fig. \ref{fig:fp_percentile} and Fig. \ref{fig:bsnn_percentile} depicts the variation of accuracy versus timesteps for different percentiles chosen from the activation histogram during threshold balancing. It is obvious that the network's latency reduces as the normalization percentile decreases due to a less conservative threshold choice. However, the accuracy degradation due to threshold relaxation seems to be minimal. Furthermore, no significant change in the number of computations are observed despite changing percentiles for both the full-precision and binary SNN models, as shown in Figs. \ref{fig:fp_perc_cmp} and \ref{fig:bsnn_perc_cmp}. The number of timesteps required to reach $62\%$ accuracy are also noted in the figure. We chose $99.7$ percentile (a subset of $3500$ training set images were used for measuring the statistics) for our remaining analysis since degradation in accuracy was observed for lower values in case of the binary model.

\begin{figure}
    \centering
    \includegraphics[scale=0.3]{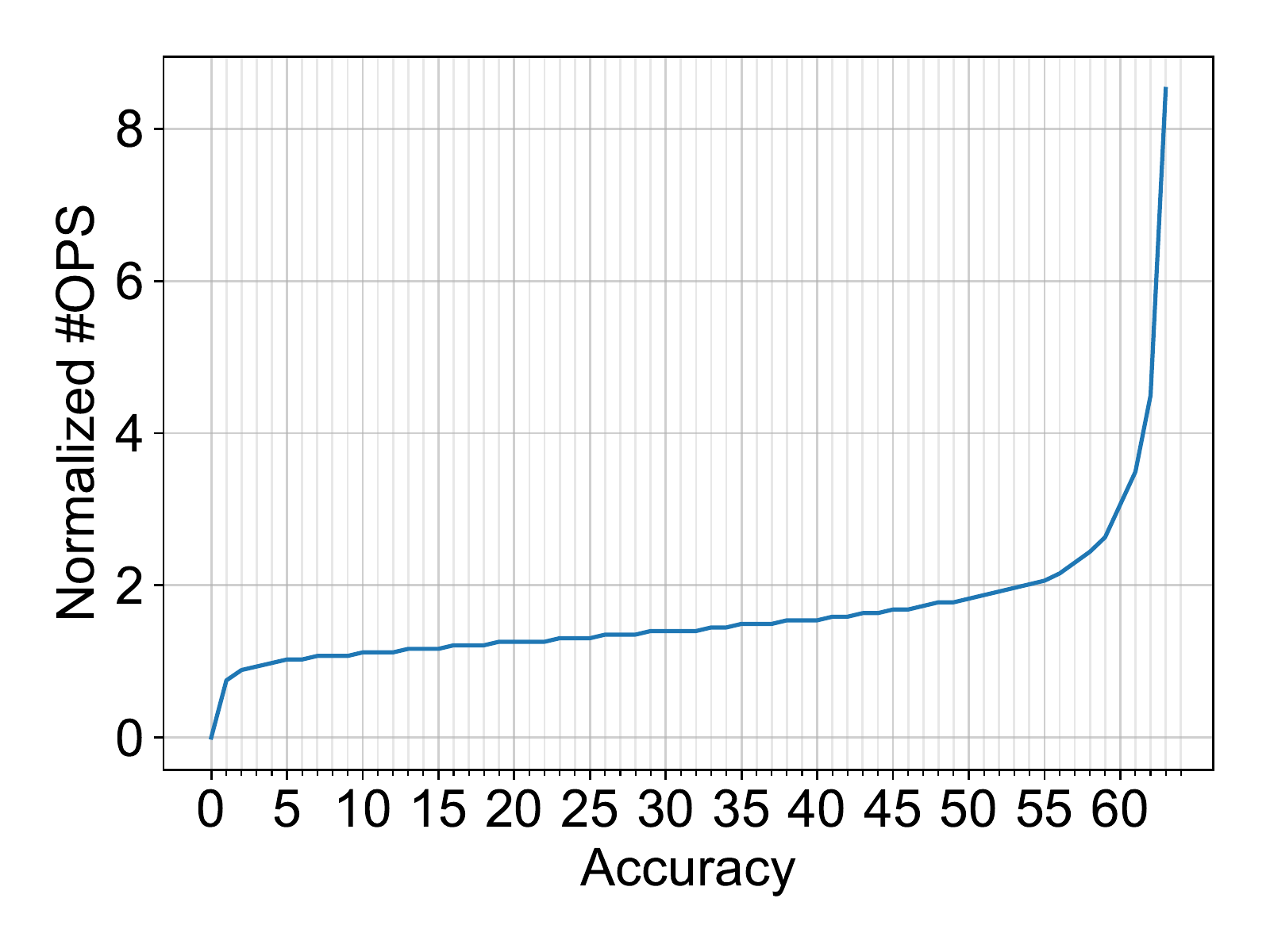}
    \caption{Normalized \#OPS of B-SNN as a function of accuracy.}
    \label{fig:thr_cmp}
\end{figure}

In order to explore additional opportunities for reduction in number of computations for the SNN models, we observed that the number of computations increases exponentially after a certain limit $\sim 60\%$ accuracy. This has been plotted in Fig. \ref{fig:thr_cmp} (combination of data shown in Figs. \ref{fig:bsnn_percentile}, \ref{fig:bsnn_perc_cmp}). Hence, computation costs for the B-SNN can be significantly reduced with small relaxation of the accuracy requirement. This is a major flexibility in our proposal unlike prior mixed-precision network proposals to circumvent the accuracy degradation issue of XNOR-Nets. The core hardware framework and operation remains almost similar to the XNOR-Net implementation with the flexibility to increase accuracy to full-precision limits as desired.

\subsubsection{Early Inference} Another conclusion obtained from the exponential increase in number of computations with accuracy beyond $\sim 60\%$ (Fig. \ref{fig:thr_cmp}) is that a few difficult images require longer evidence integration for the SNN. However, it is unnecessary to run the SNN for an extended period of time for easy image instances that could have been classified earlier. Driven by this observation, we explored an ``Early Exit" inference method for SNNs wherein we consider the SNN inference to be completed when the maximum membrane potential of the neurons in the final layer is above a given threshold\footnote{It is worth noting here the final neuron layer does not have any intrinsic membrane potential threshold, i.e. the membrane potential accumulates over time. Normal inference involves determination of the neuron with maximum membrane potential.}. This results in a dynamic SNN inference process where easier instances resulting in faster evidence integration can be classified earlier, thereby reducing the average inference latency and, in turn, the number of unnecessary computations.   

\begin{figure}[htp]
  \subfigure[The accuracy reaches $62\%$ at around voltage of $48$ and reaches $63\%$ at around voltage of $106$.]{\includegraphics[scale=0.15]{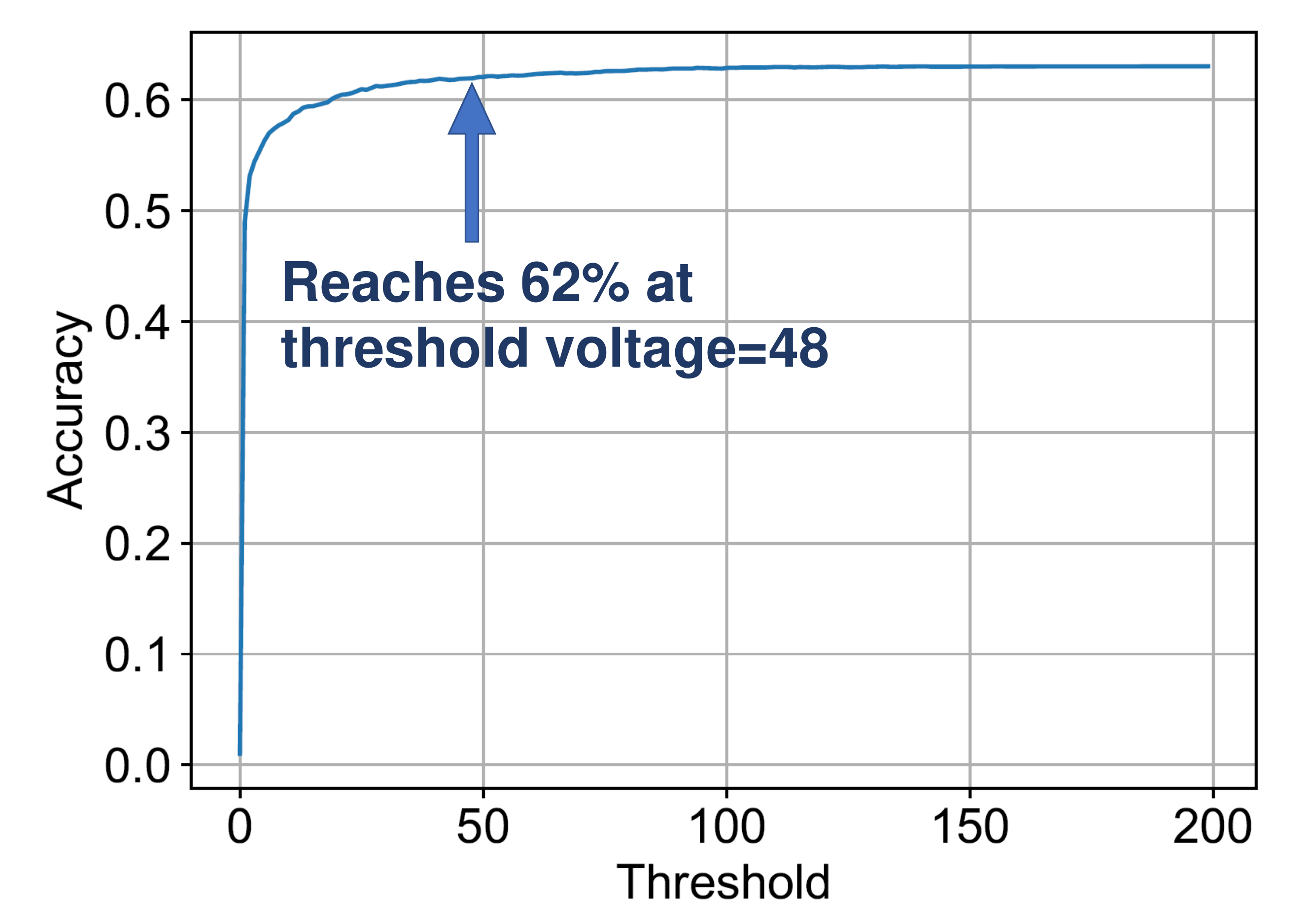}\label{fig:acc_thr}}\quad
    \subfigure[Fig. (a) at finer granularity. It reaches XNOR accuracy at threshold value $0.90$.]{\includegraphics[scale=0.15]{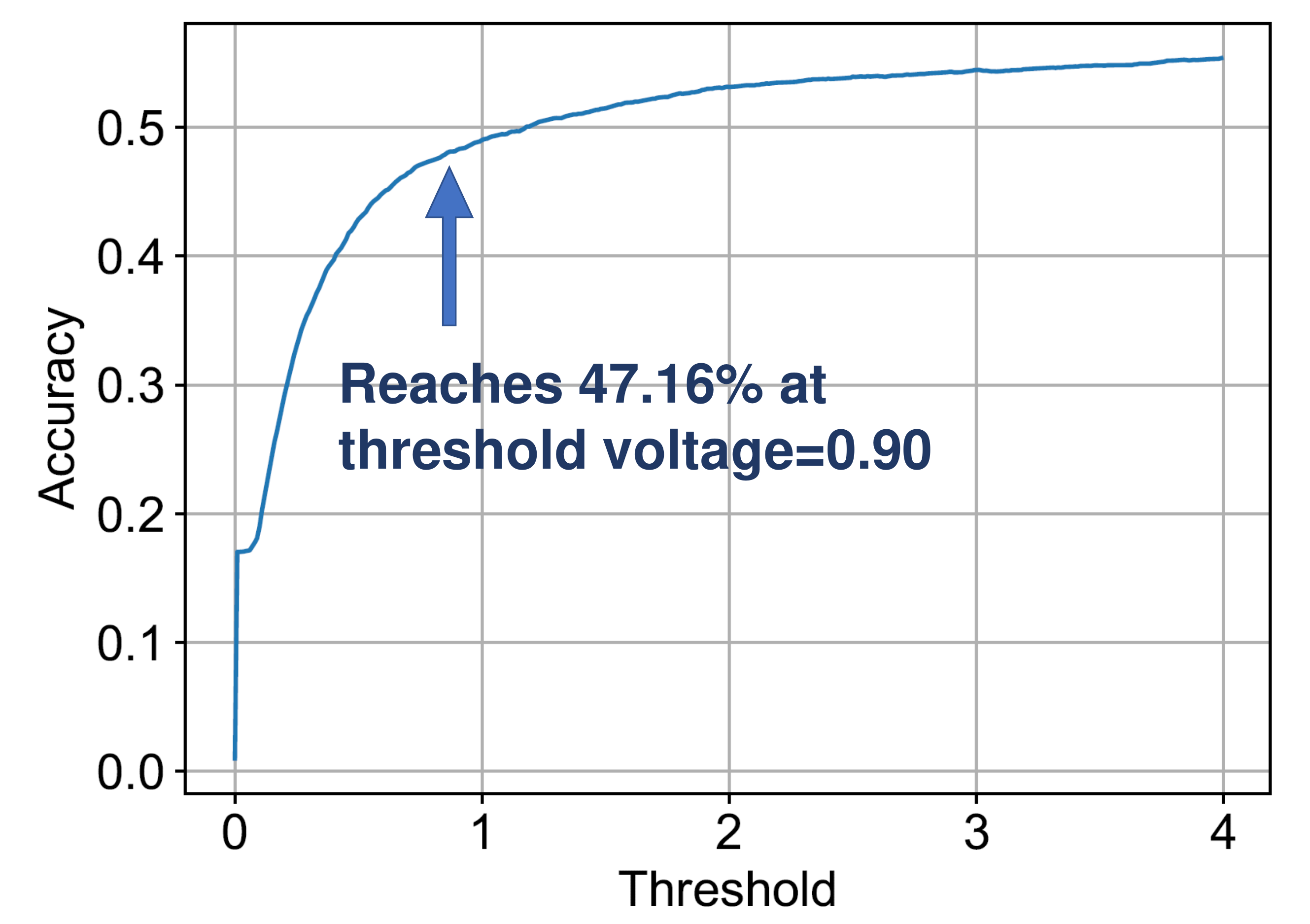}\label{fig:acc_thr_fine}}
\caption{Analysis for Early Inference.}
\end{figure}

\begin{figure}[htp]
  \centering
  \includegraphics[scale=0.3]{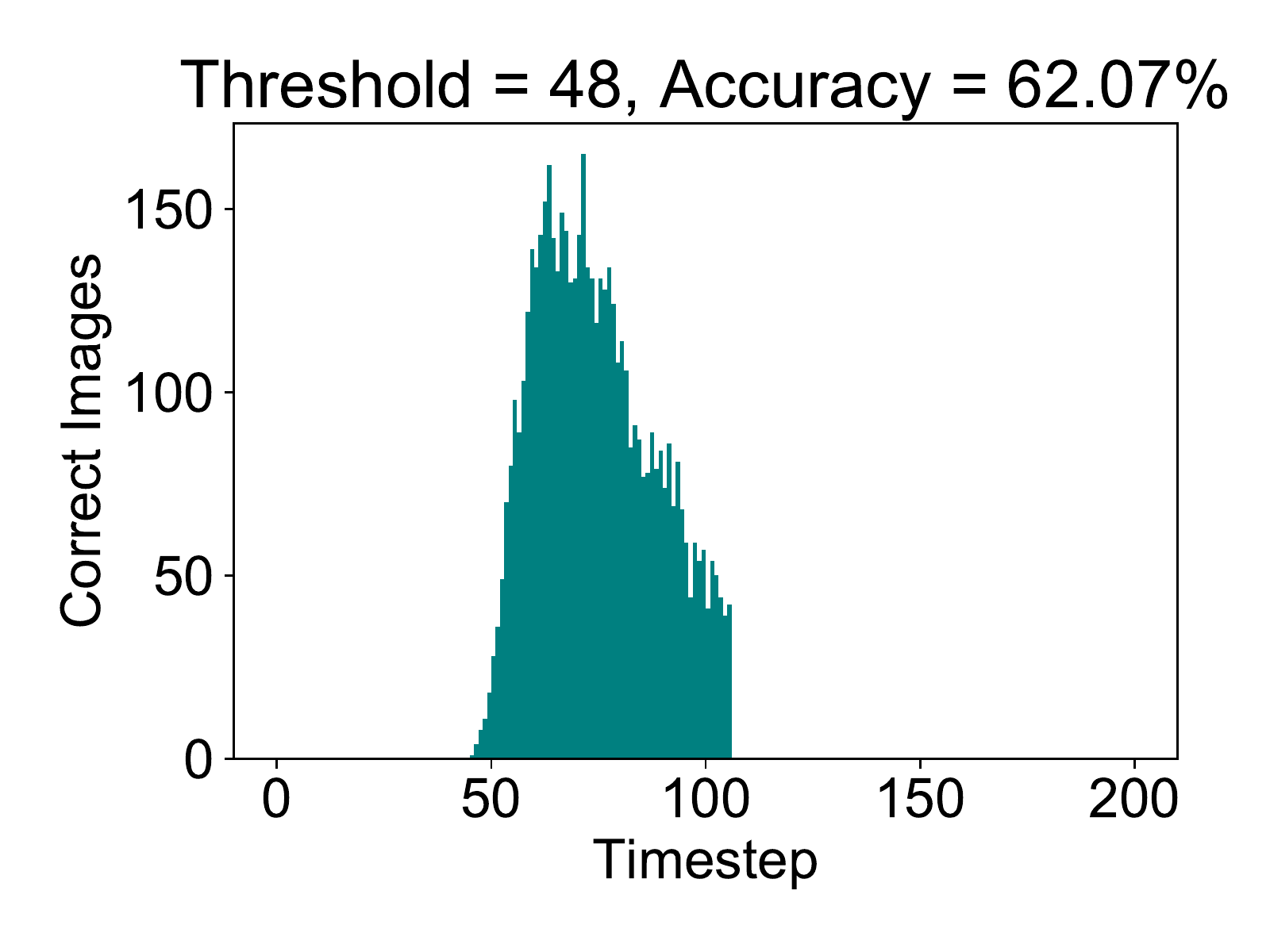}
  \caption{Evaluation with Early Inference. Note that the there are 803 images predicted at the $105$-th timestep which is not included in the graph. }\label{fig:correct_imgs}
\end{figure}

Figs. \ref{fig:acc_thr}-\ref{fig:acc_thr_fine} depicts the variation of final SNN accuracy with the confidence threshold value for the maximum membrane potential of the final B-SNN layer. This optimization is equally applicable for the full-precision model. We considered that in the worst case the SNN runs for $105$ timesteps (time required to reach baseline accuracy of $62\%$ - obtained from Fig. \ref{fig:bsnn_percentile}). Indeed, we observed a reduction in computation from \textbf{4.30} to \textbf{3.55} with early inference without any compromise in accuracy ($62\%$) for the binary model. The histogram of the required inference timesteps is shown in Fig. \ref{fig:correct_imgs}. The average value of inference timesteps is \textbf{62.4}, which is significantly lower than \textbf{105} for the case without early exit. As a comparison point, we can achieve the XNOR-Net accuracy ($47.16\%$) with threshold value $0.90$ as shown in Fig. \ref{fig:acc_thr_fine}, and the corresponding number of normalized computations is $1.49$ as compared to that of $1.0$ of XNOR. Note that the $50\%$ increment in computations for the XNOR-Net accuracy is a result of the fact that our model was optimized for a baseline accuracy of $62\%$. Hence, relaxing constraints explained in the previous subsections can potentially be used for the B-SNN to achieve XNOR-Net level accuracy at iso-computations. The results for CIFAR-$100$ dataset are summarized in Table I. The B-SNN VGG model achieves near full-precision accuracy while only requiring $3.55\times$ more operations integrated over the entire inference time window.

\begin{table}[t]
\renewcommand{\arraystretch}{1}
\caption{Results for CIFAR-$100$ Dataset}
\label{table_1}
\centering
\begin{tabular}{ p{2.8cm} p{1.1cm} p{1.4cm} }
\hline 
\hline
\bfseries {Network Model} & \bfseries {Accuracy} & \bfseries {Normalized \#OPS}\\
\hline
{Full Precision ANN} & {$64.9\%$} & {$32$} \\ \\
{XNOR Net} & {$47.16\%$} & {$1$} \\ \\
{B-SNN} & {\textbf{$62.07\%$}} & {$3.55$} \\
\hline
\hline
\end{tabular}
\end{table}

\subsection{ImageNet Results}
The full-precision VGG-$15$ model is trained on ImageNet dataset for $100$ epochs with a batch size of $128$, a learning rate of $1e-2$, a weight decay of $1e-4$ and the SGD optimizer with a momentum of $0.9$. Note that the learning rate was divided by $10$ at $30$, $60$ and $90$ epochs similar to that of CIFAR-$100$ training. The final top-1 accuracy of the full-precision ANN was $69.05\%$.

Similarly, we binarized the network from the pre-trained ANN using the hybrid methodology described previously and we also observed a drastic increase in B-SNN accuracy (in contrast to training the model from scratch) similar to Fig \ref{fig:avg_train}. The initial parameters used for the Adam optimizer were learning rate of $5e-4$, weight decay of $5e-4$ (and $0$ after 30 epochs), and beta values (the decay rates of the exponential moving averages) of ($0.0$, $0.999$). Note that we observed proper setting of the beta values to be crucial for higher accuracy of the B-SNN training, as suggested in a recent work \cite{alizadeh2018empirical}. We achieved $65.4\%$ top-1 accuracy for the constrained BWN model after 40 epochs of training (the binarization phase after full-precision training).

\begin{figure}[htp]
  \centering
  \subfigure[Accuracy versus timesteps for full-precision model.]{\includegraphics[scale=0.25]{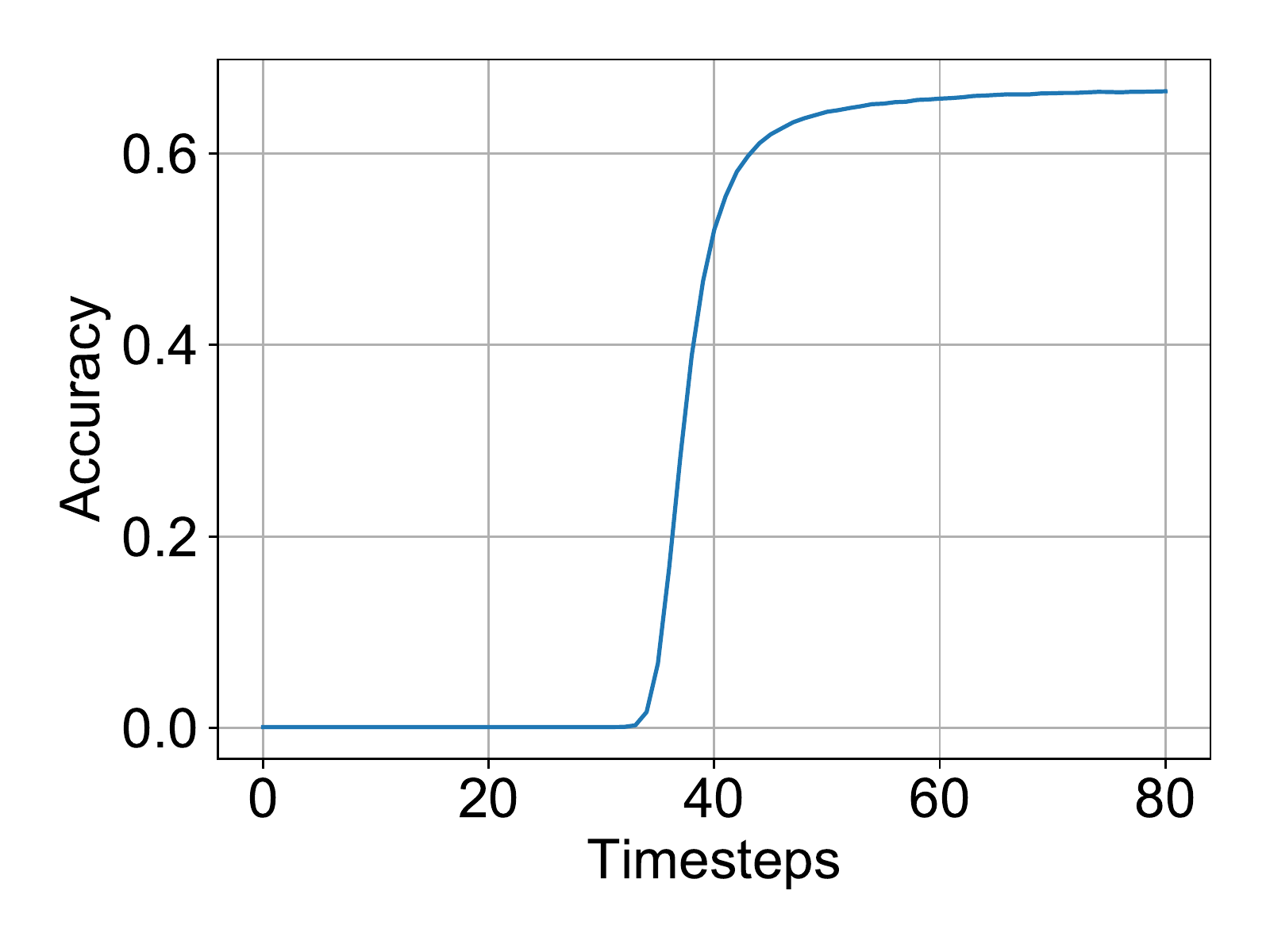}\label{fig:snn_in}}\quad
  \subfigure[Accuracy versus timesteps for binary model.]{\includegraphics[scale=0.25]{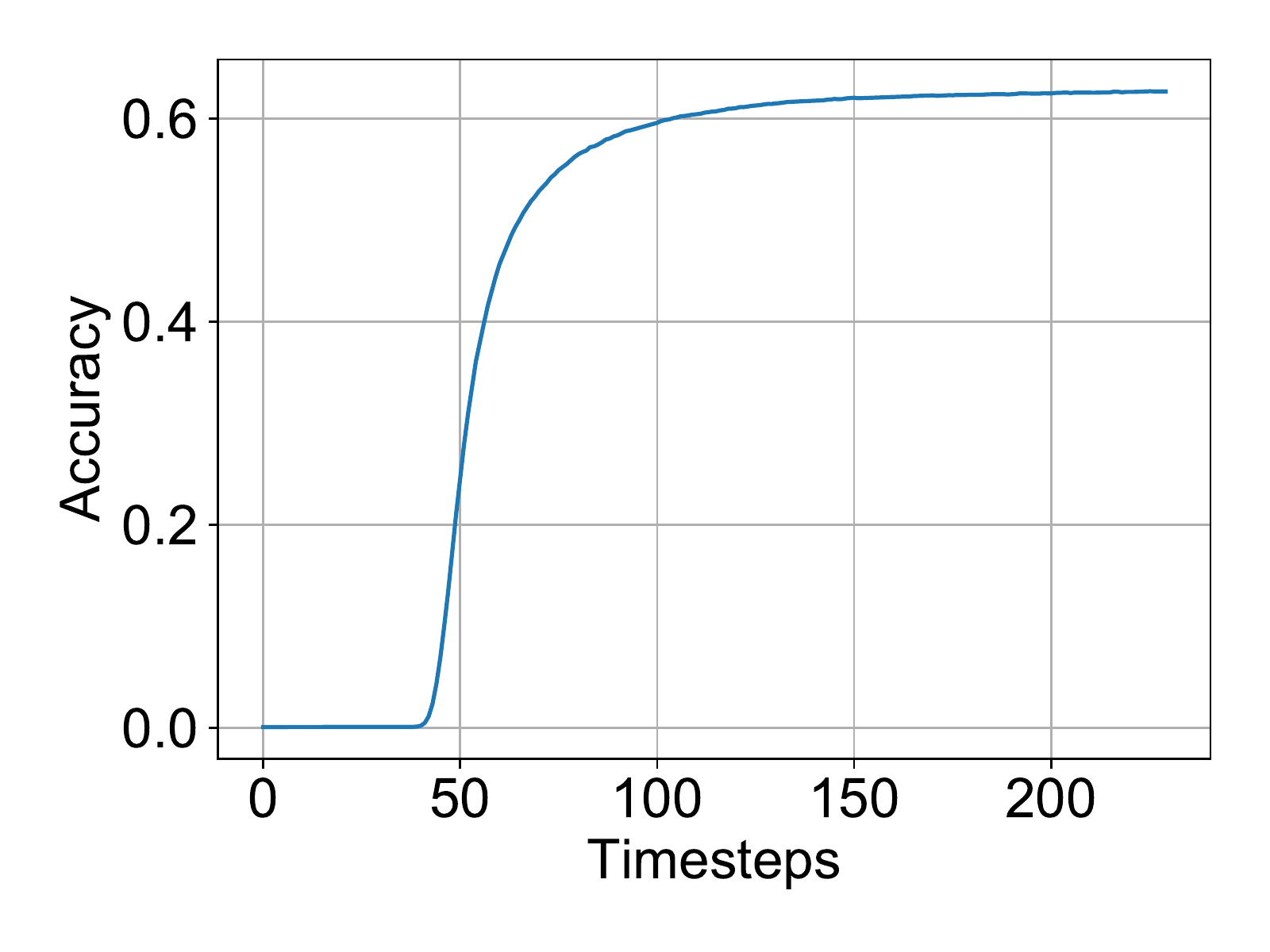}\label{fig:bsnn_in}}
  \caption{Performance on the ImageNet dataset.}
\end{figure}

Optimization settings derived from the previous CIFAR-$100$ experiments were applied to our ImageNet analysis, namely, the pooling architecture, neural node type and relaxing the threshold balancing ($99.9\%$ percentile was used by recording maximum ANN activations for a subset of $80$ images from the training set). The top-1 SNN (ANN) accuracy was $66.56\%$ ($69.05\%$)  for the full-precision model and $62.71\%$ ($65.4\%$) for the binarized model respectively. The accuracy versus timesteps variation for the two models are depicted in Fig. \ref{fig:snn_in}-(b). The binary SNN model achieves near full-precision accuracy on the ImageNet dataset as well with $5.09$ Normalized \#OPS count. Note that the latency and \#OPS count can be further reduced by early exit. We did not include the early exit optimization in order to achieve a fair comparison with previous works. A summary of our results on the ImageNet dataset and results from other competing approaches are shown in Table II. Apart from the B-SNN proposal, our simple optimization procedures involving standard non-spiking network based training is able to achieve extremely low-latency deep SNNs.
\begin{table}[h]
\renewcommand{\arraystretch}{1}
\caption{Results for ImageNet Dataset}
\label{table_1}
\centering
\begin{tabular}{ p{3.4cm} p{1.1cm} p{1.4cm} }
\hline 
\hline
\bfseries {Network Model} & \bfseries {Accuracy} & \bfseries {Timesteps}\\
\hline
{Full Precision ANN } & {$69.05\%$} & {$-$} \\ \\
{XNOR Net } & {$49.77\%$} & {$-$} \\ \\
{Full Precision SNN \newline (ANN-SNN conversion \cite{rathi2020enabling})} & {$62.73\%$} & {$250$} \\ \\
{Full Precision SNN \newline (Hybrid Training \cite{rathi2020enabling})} & {$65.19\%$} & {$250$} \\ \\
{\textbf{Full Precision SNN \newline (This Work)}} & {\textbf{66.56\%}} & \textbf{{64}} \\ \\
{\textbf{B-SNN \newline (This Work)}} & {\textbf{62.71\%}} & {\textbf{148}} \\ 
\hline
\hline
\end{tabular}
\end{table}

\section{Conclusions and Future Work}
While most of the current efforts at solving the accuracy degradation issue of BNNs have been focused on mixed-precision networks, we explore an alternative time-domain encoding procedure by exploring synergies with SNNs. ANN-SNN conversion provides a mathematical formulation for expressing multi-bit precision of ANN activations as binary values over time. Our binary SNN models achieve near full-precision accuracies on large-scale image recognition datasets, while utilizing similar hardware backbone of BNN-catered ``In-Memory" computing platforms. Further, we explore several design-time and run-time optimizations and perform extensive empirical analysis to demonstrate high-accuracy and low-latency SNNs through ANN-SNN conversion techniques. Future work will explore algorithms to reduce the accuracy gap between full-precision and binary SNNs even further along with substantiating the generalizability of the proposal to advanced network architectures like residual connections (that may require additional design considerations \cite{sengupta2019going}). Further, hardware benefits of the B-SNN proposal against mixed/reduced-precision implementations will be evaluated.

\section{Acknowledgements}
The work was supported in part by the National Science Foundation.

\end{document}